\documentclass{article} 
\usepackage{colm2024_conference}

\usepackage[utf8]{inputenc}
\usepackage[T1]{fontenc}
\usepackage{microtype}
\usepackage{hyperref}
\usepackage{url}
\usepackage{booktabs}
\usepackage{fontawesome}
\usepackage{markdown}
\usepackage{arydshln}
\usepackage{wrapfig}
\usepackage{latexsym}
\usepackage{longtable}
\usepackage{amsmath}
\usepackage{fancyvrb}
\usepackage{xcolor}
\usepackage{arydshln}
\usepackage{bm}
\usepackage{amsfonts}
\usepackage{amssymb}
\usepackage{booktabs}
\usepackage{subcaption}
\usepackage{caption}
\usepackage{graphicx}
\usepackage{multirow}
\usepackage{multicol}
\usepackage{fvextra}
\usepackage{makecell}
\usepackage{xcolor}
\usepackage{bbding}
\usepackage{amstext}
\usepackage{amsthm}
\usepackage{bbm}
\usepackage{changepage}
\usepackage{tablefootnote}

\title{RoT: Enhancing Large Language Models with \\ Reflection on Search Trees}

\author{Wenyang Hui, ~~Kewei Tu\\
School of Information Science and Technology, ShanghaiTech \\
Shanghai Engineering Research Center of Intelligent Vision and Imaging\\
\texttt{\{huiwy, tukw\}@shanghaitech.edu.cn},
}

\colmfinalcopy 
\begin{document}

\maketitle

\begin{abstract}
Large language models (LLMs) have demonstrated impressive capability in reasoning and planning when integrated with tree-search-based prompting methods. However, since these methods ignore the previous search experiences, they often make the same mistakes in the search process. To address this issue, we introduce \textbf{Reflection on search Trees (RoT)}, an LLM reflection framework designed to improve the performance of tree-search-based prompting methods. It uses a strong LLM to summarize guidelines from previous tree search experiences to enhance the ability of a weak LLM. The guidelines are instructions about solving this task through tree search which can prevent the weak LLMs from making similar mistakes in the past search process. In addition, we proposed a novel state selection method, which identifies the critical information from historical search processes to help RoT generate more specific and meaningful guidelines. In our extensive experiments, we find that RoT significantly improves the performance of LLMs in reasoning or planning tasks with various tree-search-based prompting methods (e.g., {\bf BFS} and {\bf MCTS}). Non-tree-search-based prompting methods such as Chain-of-Thought ({\bf CoT}) can also benefit from RoT guidelines since RoT can provide task-specific knowledge collected from the search experience. The code is available at \url{https://github.com/huiwy/reflection-on-trees}.

\end{abstract}
\section{Introduction}
Recent research highlighted that tree-search-based prompting methods significantly improve the models' capability of reasoning and planning in tasks that require multiple steps of formal reasoning or planning, such as embodied planning \citep{rap}, mathematical reasoning \citep{tot} and dialogue policy planning \citep{mcts-dialogue}. 

\begin{figure}[t]
    \centering
    \includegraphics[width=0.8\linewidth]{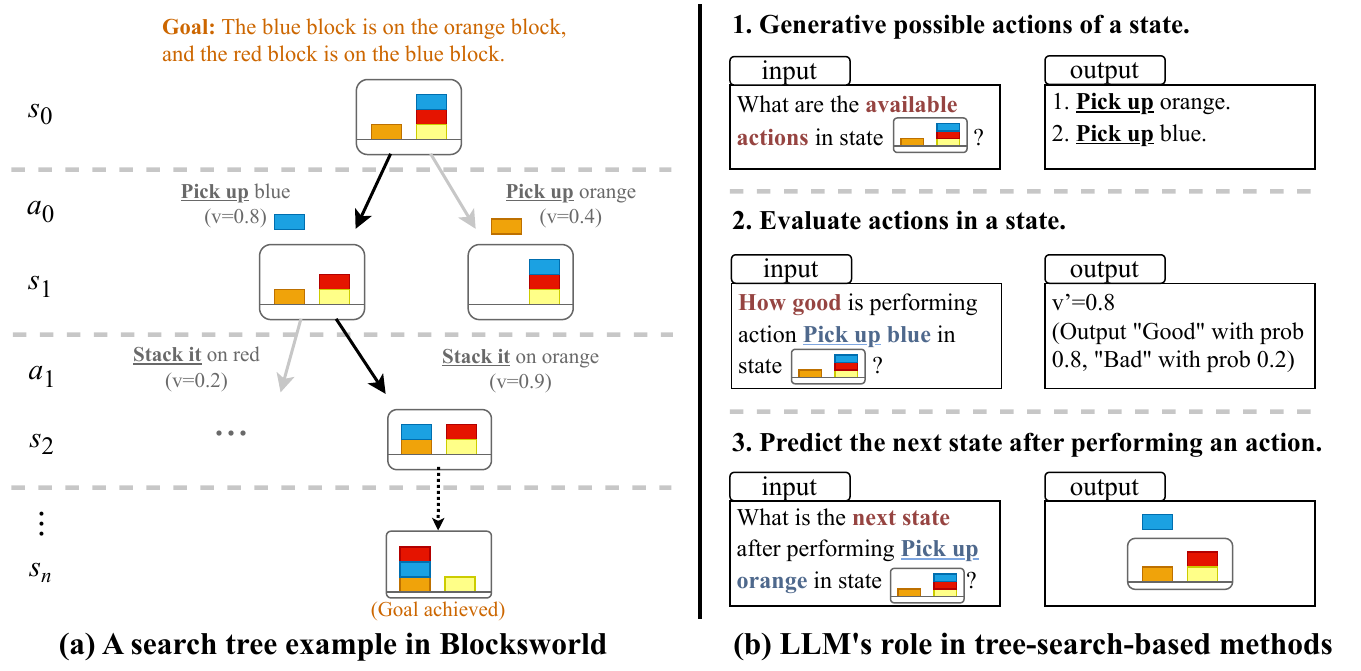}
    \caption{An illustration about tree-search-based prompting method in \textbf{Blocksworld}. $a_i$ and $s_i$ denotes action and state at depth $i$. $v$ is the estimated value of an action by the tree search algorithm (value estimation in BFS, and average estimated value of children in MCTS).}
    \label{fig:search-tree-intro}
\end{figure}

These methods decompose the problem into multiple steps to solve each step sequentially. In each step they try an {\bf action}, leading to a {\bf state} transition, gradually approaching the correct answer. For instance, as depicted in Figure \ref{fig:search-tree-intro}(a), \textbf{Blocksworld} \citep{blocksworld} is a task to manipulate the blocks from the initial state to the goal state. Following RAP's \citep{rap} formulation of \textbf{Blocksworld}, the actions are ways to manipulate the blocks and states are the current block configuration. Then, they use tree search methods such as breadth-first search (BFS) to search for an optimal path of states and actions that leads to the solution, with the assistance of LLMs. They employ LLMs to generate available actions, predict states after applying an action, and evaluate the states or actions (Figure \ref{fig:tree-search-example}). 

However, since these methods do not learn from past failures, they often make repeated mistakes, including incorrectly evaluating the actions, generating actions leading to low outcomes, and failing to predict the next state. These issues result in low accuracy and poor search efficiency, leading to the over-exploration of the wrong actions \citep{treedisc}. To empower tree-search-based prompting methods to learn from their search experiences, we introduce \textbf{Refection on search Trees (RoT)}, a framework designed to improve search efficiency and accuracy by reflecting on previous tree search experiences. RoT employs a strong LLM to reflect on the previous search process of a weak LLM to get a natural language task-level guideline. This guideline is then used to enhance the weak LLM's capability of making the right decisions and estimations during the subsequent search processes. To generate a better guideline, we designed a critical information extraction mechanism to select the most crucial states that significantly impact the outcomes from the previous search trees. With guidelines, models can avoid repeating mistakes and make better decisions. 

We evaluated RoT on a variety of complex reasoning and planning tasks including embodied planning in \textbf{Blocksworld} \citep{blocksworld}, mathematical reasoning in \textbf{GSM8k} \citep{gsm8k}, and dialogue policy planning in \textbf{CraigsListBargain} \citep{bargain}. 
In our experiments, RoT significantly improves the performance of various strong LLMs in these tasks when using tree-search-based prompting methods such as BFS and MCTS. RoT also outperforms the recently proposed strong reflection method LEAP \cite{leap-icl} designed for non-tree-search-based methods especially when the problem is hard.
Non-tree-search-based prompting methods such as Chain-of-Thought (CoT) \citep{cot,cot2} and Self-Consistency \citep{self-consistency} can also benefit from the guidelines generated by RoT since they can provide task knowledge collected from the search experience. We also find that among all the tasks, RoT has the greatest benefit to tasks that models are not familiar with.

\section{Background}

\subsection{Tree Search Methods}
Tree search methods such as BFS, A* search \citep{astar}, and Monte Carlo Tree Search \citep{mcts1,mcts2} are used to search a tree for an optimal state. A search tree $\mathcal{T}$ consists of states $s$, actions $a$, state value estimations $V$, and actions value estimations $Q$. Actions are transitions between states. Each action within a state is associated with a value $Q(s,a)$ which estimates the quality of action $a$ in state $s$. Additionally, each state is assigned a value $V(s)$ estimating the quality of state $s$. Since in the deterministic setting, the incoming action of a state is unique, we set $V(s')$ to $Q(s,a)$ if taking action $a$ in state $s$ transitions to $s'$. Consequently, the primary objective of tree search methods is to identify a sequence of actions $\textbf{a}$ that leads to the state $s^*$ with the highest value $V(s^*)$.

\subsection{Tree-Search-Based Prompting Methods}
Integrating tree-search methods with LLMs presents an intriguing approach to identifying a series of reasoning steps leading to the correct solution. As depicted in Figure \ref{fig:search-tree-intro}(b), throughout the tree search process, the LLM can be prompted to (1) generate the available actions $\{a\}$ within a state $s$, (2) assess the value $Q'(s,a)$ of taking action $a$ within state $s$, and (3) predict the next state $s'$ after taking action $a$ in state $s$. With insights from the LLM, tree-search methods can find an effective sequence of actions leading to a favorable outcome.

\begin{figure}
    \centering
    \includegraphics[width=0.9\linewidth]{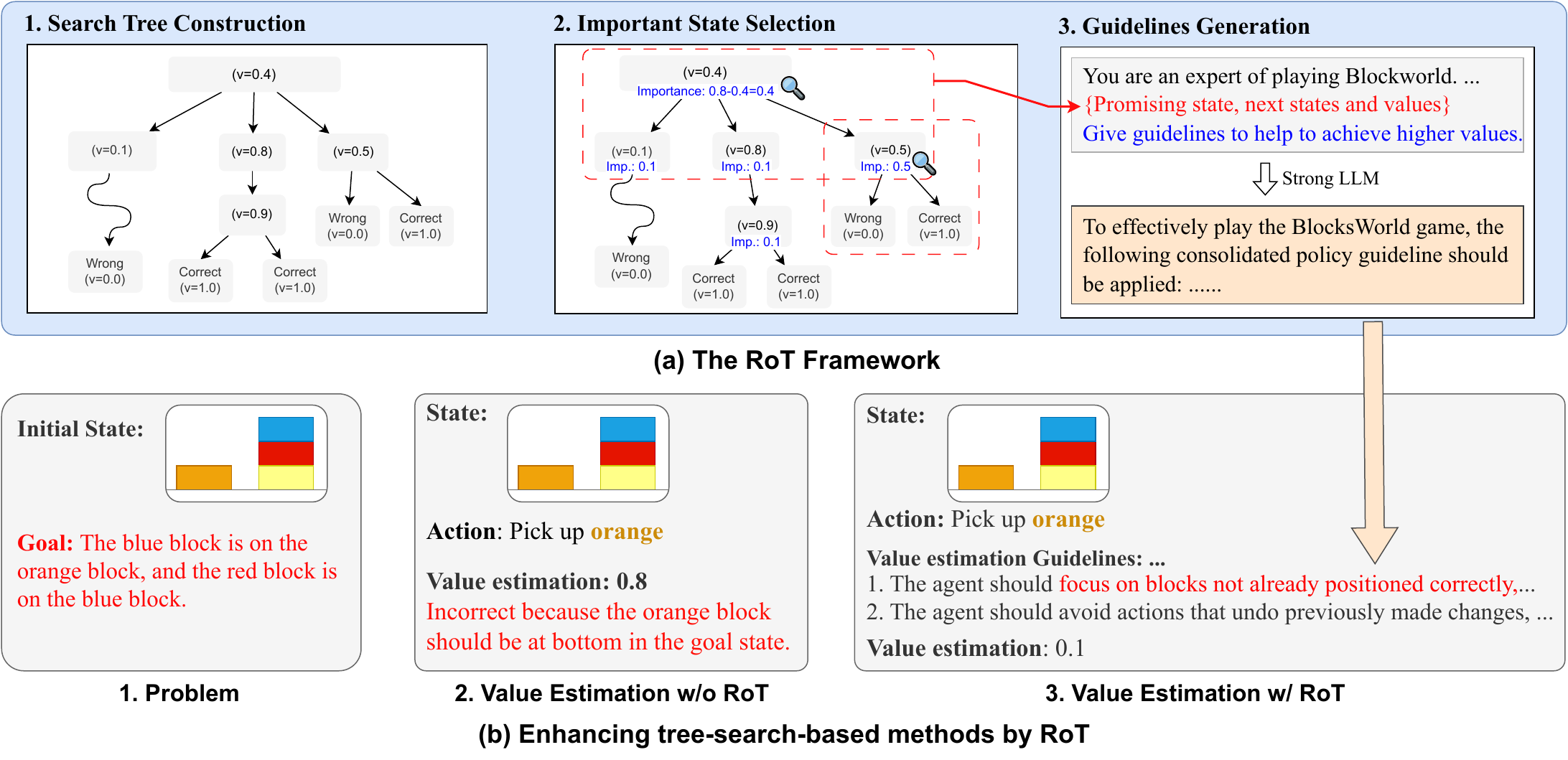}
    \caption{The RoT framework}
    \label{fig:framework}
\end{figure}

\section{Reflection on Search Trees}
The framework of RoT is depicted in Figure \ref{fig:framework}. We use tree-search-based prompting methods to construct the search trees. Then, as explained in Section \ref{sec:sel} the important states where making wise decisions can greatly improve future outcomes are selected from the generated search trees. Finally, the guidelines are summarized from these decisions to improve the future search process according to Section \ref{sec:reflection}.

\subsection{Important State Selection from the Search Tree} \label{sec:sel}
\begin{wrapfigure}{r}{0.49\textwidth}
    \centering
    \includegraphics[width=0.49\textwidth,trim=0 0 40 0,clip]{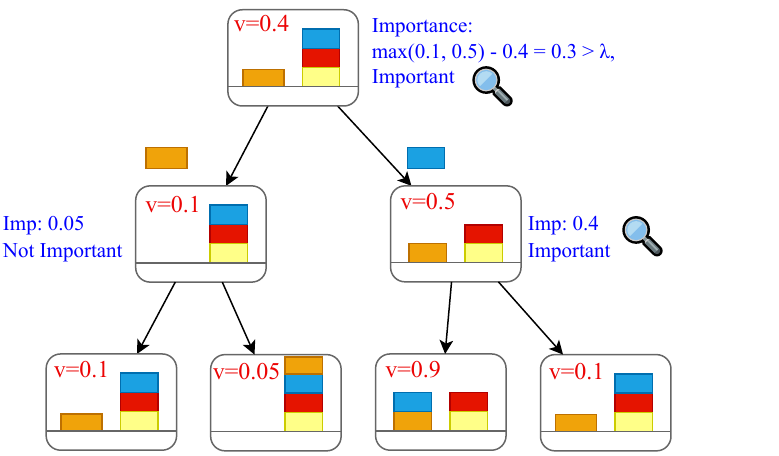}
    \caption{Important State Selection ($\lambda=0.1$). States marked \includegraphics[width=0.04\textwidth, trim=2 8 2 2]{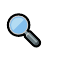}are selected.}
    \label{fig:promising}
\end{wrapfigure}

We extract the most informative experiences from previous tree search experiences by important state selection. A state is important if taking an action $a$ within state $s$ leads to a great increase or decrease in the state value $V(s')$. Intuitively, making wise decisions within such states can vastly improve tree search performance. Then through reflection on such experiences, the model can learn to make better decisions and attain better performance. Specifically, as depicted in Figure \ref{fig:promising}, given the search tree $\mathcal{T}$, we calculate the importance score of state $s$ based on the values of their children in the following way:  $$\text{Importance}(s)=\max_{s'\in children(s)}|V(s')-V(s)|$$
Then, a state $s$ is called important when $\text{Importance}(s)>\lambda$, where $\lambda$ is a threshold.
RoT also saves relevant information in the search tree about the important states for better reflection. Specifically, apart from the important state itself, RoT collects information about the available actions, the next state corresponding to each action, and the values of the next states.

\subsection{Guideline Summarization} \label{sec:reflection}

\begin{figure}[t]
    \centering
    \includegraphics[width=0.9\linewidth,trim=0 0 0 0,clip]{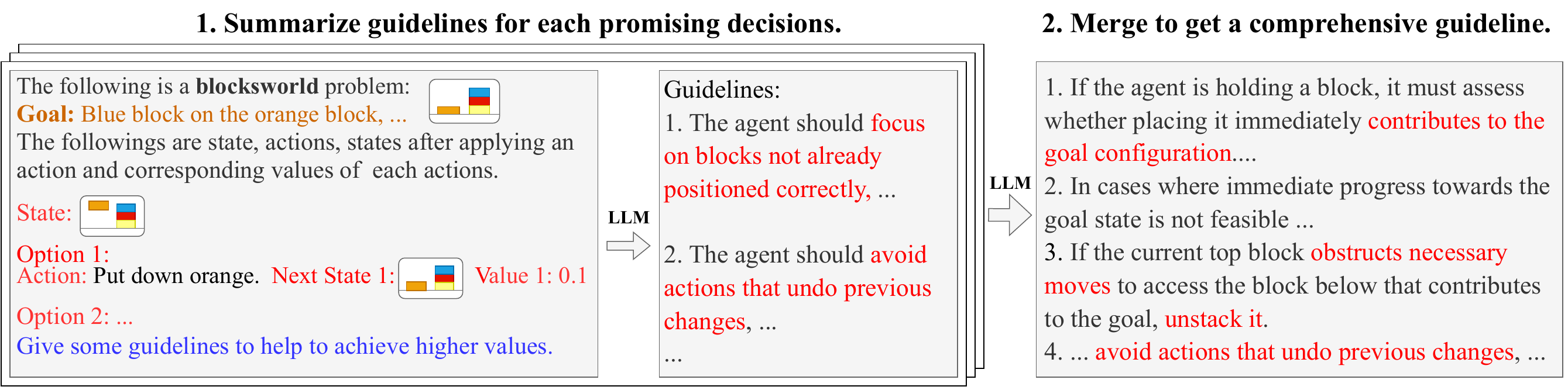}
    \caption{Guideline Summarization.}
    \label{fig:reflection}
\end{figure}

Based on the information collected in important state selection, RoT summarizes a guideline to enhance future search performance with the assistance of LLM.  As described in Figure \ref{fig:reflection}.1, for each important state, we ask a strong LLM to contrastively reflect on all the actions and the consequences of taking action. We prompt this LLM first to analyze the impact of each action on the value of the next states, and then summarize a guideline based on its analysis. Finally, the weaker LLM is given this guideline to improve the tree search efficiency. 

However, when presenting LLMs with information about all the important states simultaneously, LLM often gets overwhelmed and fails to capture the details in summarization, resulting in overly general guidelines. Consequently, we separately generate guidelines from a single important state and then prompt LLM to merge the guidelines into one comprehensive guideline (Figure \ref{fig:reflection}.2). 
To improve the search efficiency and quality in the future search process, RoT appends the generated guidelines to a predefined location in the prompt used to generate action, states, or estimate values. With the enhanced prompts, LLMs can attain better performance in tree searches. Beyond enhancing search performance, the summarized guidelines can augment non-tree-search-based methods such as Chain-of-Thought (CoT) by providing LLMs with prior knowledge about the task.

\subsection{Iterative Improvement} \label{sec:iterative}
Inspired by expert iteration \citep{expertiter}, an approach to further enhance search performance is applying the RoT iteratively. This uses previously generated guidelines to generate improved search trees and summarizes an enhanced guideline based on the improved search tree and past guidelines to prevent the mistakes that are hard to avoid. 
\begin{figure}[b]
    \centering
    \includegraphics[width=0.9\linewidth]{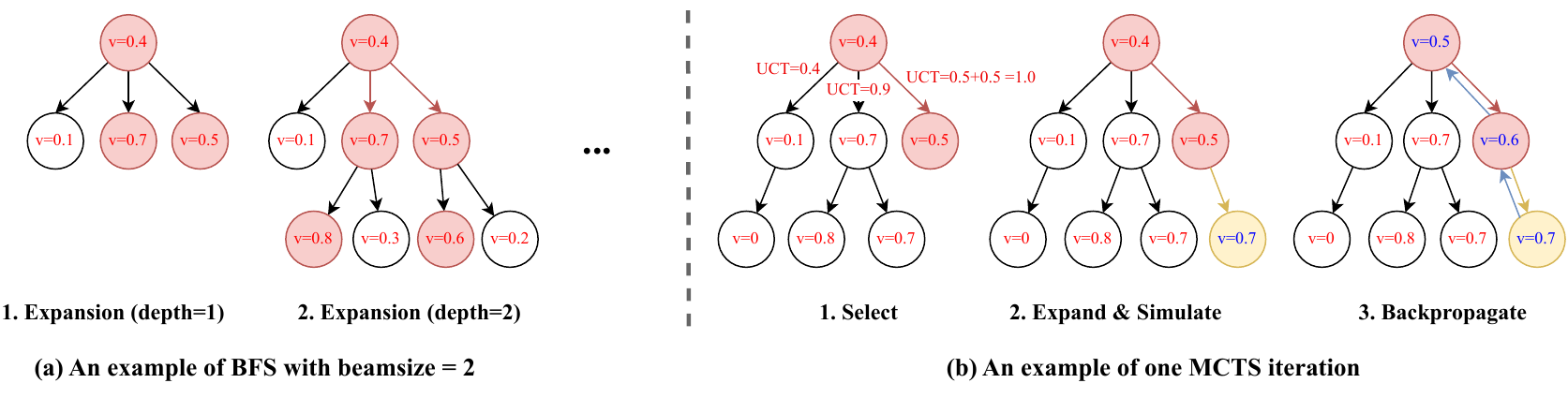}
    \caption{Examples of BFS and MCTS.}
    \label{fig:tree-search-example}
\end{figure}

\section{Experiments}
In this section, we evaluate the effectiveness of RoT on complex reasoning and planning tasks by integrating it with two tree-search-based methods, breadth-first search (BFS) and (Monte Carlo Tree Search) MCTS \cite{mcts1}, and two non-tree-search-based methods, Chain of Thoughts (CoT) \citep{cot} and CoT with self-consistency \citep{self-consistency}. We focus on analyzing the performance improvement on these prompting methods after applying RoT summarized guidelines. Meanwhile, we also compare RoT with the recent non-tree-search-based reflection method, LEAP \citep{leap-icl}. 
\begin{table}[t]

    \centering
    \scalebox{0.63}{
    \begin{tabular}{ll|cc|c||cc|c||cc|c||cc|c}
    \toprule
     & \multirow{2}[3]{*}{Method} & \multicolumn{3}{c||}{Step 4}  & \multicolumn{3}{c||}{Step 6}  & \multicolumn{3}{c||}{Step 8}  & \multicolumn{3}{c}{Step 10}  \\ \rule{0pt}{11px}
    && Base & \bf  RoT & {LEAP} &  Base & \bf  RoT & {LEAP} &  Base & \bf  RoT & {LEAP} &  Base & \bf  RoT & {LEAP} \\ \midrule
    \multirow{5}{*}{\rotatebox{90}{\textbf{phi-2\ \ \ \ }}} & CoT & 1.3 & \textbf{3.9} {\scriptsize {\color{red}(+200.0\%)}} & {1.7} & 0.0 & 0.0 {\scriptsize {\color{black}(+0.0\%)}} & {0.0} & 0.0 & 0.0 {\scriptsize {\color{black}(+0.0\%)}} & {0.0} & 0.0 & 0.0 {\scriptsize {\color{black}(+0.0\%)}} & {0.0}\\
 & CoT-Pass@10 & 5.3 & {\bf{15.8}} {\scriptsize {\color{red}(+198.1\%)}} & {13.2} & 1.4 & 1.4 {\scriptsize {\color{black}(+0.0\%)}} & \bf{2.1} & 0.0 & \textbf{0.7} {\scriptsize {\color{black}(+0.0\%)}} & {0.0} & 0.0 & 0.0 {\scriptsize {\color{black}(+0.0\%)}} & {0.0}\\
 & BFS$^{(5)}$ & 61.8 & \textbf{69.7} {\scriptsize {\color{red}(+12.8\%)}} & {67.1} & 25.5 & {29.0} {\scriptsize {\color{red}(+13.7\%)}} & \bf{33.1} & 11.2 & \textbf{16.1} {\scriptsize {\color{red}(+43.8\%)}} & 14.0 & 5.8 & \textbf{8.7} {\scriptsize {\color{red}(+50.0\%)}} & {5.8}\\
 & MCTS$^{(1)}$& 30.3 & \textbf{36.8} {\scriptsize {\color{red}(+21.5\%)}} & {28.9} & 17.9 & {18.6} {\scriptsize {\color{red}(+3.9\%)}} & \bf{22.8} & 4.2 & \textbf{6.3} {\scriptsize {\color{red}(+50.0\%)}} & {4.2} & \textbf{1.9} & \textbf{1.9} {\scriptsize {\color{black}(+0.0\%)}} & \textbf{1.9}\\
 & MCTS$^{(10)}$&84.2 & \textbf{85.5} {\scriptsize {\color{red}(+1.5\%)}} & {80.3} & 46.9 & \textbf{55.2} {\scriptsize {\color{red}(+17.7\%)}} & {53.1} & 11.2 & \textbf{17.5} {\scriptsize {\color{red}(+56.2\%)}} & {14.7} & 2.9 & \textbf{3.9} {\scriptsize {\color{red}(+34.5\%)}} & {2.9}\\
\midrule
\multirow{5}{*}{\rotatebox{90}{\textbf{mistral-7b\ \ \ }}} & CoT & 5.3 & {10.5} {\scriptsize {\color{red}(+98.1\%)}} & \bf {10.9} & 3.4 & \textbf{6.2} {\scriptsize {\color{red}(+82.4\%)}} & {3.8} & 0.0 & 0.0 {\scriptsize {\color{black}(+0.0\%)}} & {0.0} & 0.0 & {1.0} {\scriptsize {\color{black}(+0.0\%)}} & \bf{1.2}\\
 & CoT-Pass@10 &40.8 & {48.7} {\scriptsize {\color{red}(+19.4\%)}} & \bf{55.3} & {24.1} & 21.4 {\scriptsize {\color{blue}(-11.2\%)}} & \bf{26.9} & 4.9 & \textbf{10.5} {\scriptsize {\color{red}(+114.3\%)}} & {9.0} & 7.8 & \textbf{11.7} {\scriptsize {\color{red}(+50.0\%)}} & {10.7}\\
 & BFS$^{(5)}$ & 86.8 & \textbf{89.5} {\scriptsize {\color{red}(+3.1\%)}} & \textbf{89.5} & {64.1} & 61.4 {\scriptsize {\color{blue}(-4.2\%)}} & \bf{62.1} & 42.0 & \textbf{42.7} {\scriptsize {\color{red}(+1.7\%)}} & {42.0} & 32.0 & \textbf{33.0} {\scriptsize {\color{red}(+3.1\%)}} & \textbf{33.0}\\
 & MCTS$^{(1)}$ &59.2 & \textbf{61.8} {\scriptsize {\color{red}(+4.4\%)}} & \textbf{61.8} & 40.0 & \textbf{43.4} {\scriptsize {\color{red}(+8.5\%)}} & {41.4} & 14.5 & \textbf{24.5} {\scriptsize {\color{red}(+69.0\%)}} & {21.0} & 12.6 & \textbf{21.4} {\scriptsize {\color{red}(+69.8\%)}} & {18.4}\\
 & MCTS$^{(10)}$&89.5 & {92.1} {\scriptsize {\color{red}(+2.9\%)}} & \bf{94.7} & {76.6} & 76.0 {\scriptsize {\color{blue}(-0.8\%)}} & \bf{77.2} & 55.2 & \textbf{59.4} {\scriptsize {\color{red}(+7.6\%)}} & {58.0} & 29.1 & \textbf{34.0} {\scriptsize {\color{red}(+16.8\%)}} & {30.1}\\
\midrule
\multirow{5}{*}{\rotatebox{90}{\textbf{mixtral-8x7b\ }}} & CoT & 9.2 & \textbf{18.4} {\scriptsize {\color{red}(+100.0\%)}} & {13.0} & 11.0 & \textbf{11.7} {\scriptsize {\color{red}(+6.4\%)}} & {9.4} & 2.1 & \textbf{4.9} {\scriptsize {\color{red}(+133.3\%)}} & {2.8} & 2.9 & \textbf{3.9} {\scriptsize {\color{red}(+34.5\%)}} & {3.3}\\
 & CoT-Pass@10 & 63.2 & \textbf{71.1} {\scriptsize {\color{red}(+12.5\%)}} & {61.8} & \bf54.5 & \textbf{54.5} {\scriptsize {\color{black}(+0.0\%)}} & {53.1} & 17.5 & \textbf{29.4} {\scriptsize {\color{red}(+68.0\%)}} & {21.7} & 20.4 & \textbf{27.2} {\scriptsize {\color{red}(+33.3\%)}} & {26.2}\\
 & BFS$^{(5)}$ & 92.1 & \textbf{93.4} {\scriptsize {\color{red}(+1.4\%)}} & {92.1} & \textbf{60.7} & 57.2 {\scriptsize {\color{blue}(-5.8\%)}} & {54.5} & 33.6 & \textbf{40.6} {\scriptsize {\color{red}(+20.8\%)}} & {39.2} & 29.1 & \textbf{33.0} {\scriptsize {\color{red}(+13.4\%)}} & {32.0}\\
 & MCTS$^{(1)}$ & 64.5 & \textbf{69.7} {\scriptsize {\color{red}(+8.1\%)}} & {67.1} & 41.9 & \textbf{42.8} {\scriptsize {\color{red}(+2.1\%)}} & {42.1} & 17.5 & \textbf{21.7} {\scriptsize {\color{red}(+24.0\%)}} & {20.3} & 16.5 & \textbf{22.3} {\scriptsize {\color{red}(+35.2\%)}} & {15.5}\\
 & MCTS$^{(10)}$  & 88.2 & \textbf{89.5} {\scriptsize {\color{red}(+7.4\%)}} & {86.8} & \textbf{76.6} & 75.2 {\scriptsize {\color{blue}(-1.8\%)}} & {75.2} & 51.7 & \textbf{54.5} {\scriptsize {\color{red}(+5.4\%)}} & {50.3} & 32.0 & \textbf{34.0} {\scriptsize {\color{red}(+6.2\%)}} & {29.1}\\

\bottomrule
    \end{tabular}
        }
    \caption{Partial \textbf{Blocksworld} Results. CoT-Pass$^{(n)}$ is the rate that at least one correct solution in $n$ CoT generations. BFS$^{(b)}$ means the number of beams in BFS is $b$. MCTS$^{(n)}$ denotes $n$ iterations are performed. Steps $k$ is the minimum number of actions to solve this problem. A higher number of steps indicates a harder problem. The full table is shown in Table \ref{tab:bw-res-full}.}
    \label{tab:bw-res}
\end{table}

We will first introduce the baseline methods used in our experiments. After that, we introduce the tasks, experiment settings, and their main results. Finally, we did ablation studies about important state selection and iterative improvement. Due to the limited space, we show the experiments about the iterative improvement of RoT in Appendix \ref{sec:iter-imp}.

\subsection{Baselines}
\noindent
\textbf{BFS} explores the search tree by depth. As depicted in Figure \ref{fig:tree-search-example}(a), BFS expands all the actions in each state in each depth to generate states of the next depth iteratively until a depth limit is reached. Since exploring all actions leads to exponential computational costs, beam BFS is adopted. Only actions in the $b$ best states are expanded, where $b$ is the number of beams. The scores of states are obtained by prompting the LLM to predict the score once. 

\noindent \textbf{MCTS} is an efficient tree search method. As shown in Figure \ref{fig:tree-search-example}(b), it first selects a promising state and action that scores the highest in exploration and exploitation. Then it takes this action to expand a new state. Then it evaluates the quality of this state and backpropagates the evaluation for a more precise quality estimation of its ancestors. The value estimations of a state and action are determined by the LLM's estimation of itself and the estimations propagated from its descendants through backpropagation.

\noindent \textbf{CoT and CoT-SC} are two widely used non-tree-search-based prompting methods. CoT requires the model to generate a reasoning chain to improve the reasoning ability. CoT-SC samples multiple reasoning paths and conducts a majority vote to select the best answer.

\noindent \textbf{LEAP} is a reflection method designed for CoT. It uses a strong LLM to generate guidelines based on comparisons between generated wrong answers and gold answers in the trainset to prevent the model from making similar mistakes in reasoning in the future.


\subsection{Blocksworld}
\textbf{Blocksworld} requires the agent to have a strong ability to plan and awareness of the future consequences. The task involves manipulating stacks of blocks into a goal state where the blocks in the stacks satisfy the desired configuration. At each step, the agent can pick up a block on top of a stack when the hand is empty, place the held block onto a stack of blocks, or put the block onto the ground. 

\noindent
\textbf{Task setup.} We follow RAP \citep{rap}'s setup in tree search and CoT. The states are natural language descriptions of block configurations, and actions are block manipulations. Rules are used to generate the actions within. After an action is taken, we use rules to get the next state. The initial value estimation of actions is computed by normalized log probabilities of each action provided by LLMs conditioned on the current state. A search tree of \textbf{Blocksworld} is in Figure \ref{fig:tree-search-example}. The sample LLM input and output are in Appendix \ref{sec:block-sample}. 

When performing RoT and LEAP, we generate 20 samples and summarize the guidelines on these samples with \textbf{gpt4} \citep{gpt4}. We perform MCTS for 20 iterations to generate search experiences and set the threshold $\lambda$ to $0.1$.

We use the minimum required steps to partition the test set. As the number of steps increases, the complexity of the samples correspondingly increases. This allows us to analyze the performance of our system on tasks with different difficulties. We evaluate RoT on three LLMs: \textbf{phi-2} \citep{phi-2}, a 2.7B model, \textbf{mistral-7b} \citep{mistral}, a 7B model, and \textbf{mixtral-8x7b} \citep{mixtral}, a 8x7B mixture of experts model. The specific model names or URLs are listed in Appendix \ref{sec:model}

\noindent
\textbf{Results.} As shown in Table \ref{tab:bw-res}, RoT significantly improves the accuracy of tree-search-based methods and non-tree-search-based methods over baseline. Notably, RoT demonstrates consistent performance improvements across a range of models. Compared to LEAP, RoT attains better performance on average, as it summarizes the guidelines based on important states. By reflecting on states, the LLM can generate more specific guidelines, while LEAP reflects on the whole solution, making the guidelines less specific and thus less helpful. By comparing the results from different steps, the relative improvement from RoT gets larger as the number of required steps gets larger. For instance, the average relative improvement to the baseline of MCTS increases from $+2.2\%$ at step 2 to $+27.1\%$ at step 8. This indicates RoT gets more effective when the task gets harder.

\subsection{GSM8k}
\textbf{GSM8k} \citep{gsm8k} is a math word problem dataset, where each problem consists of a description and a final question asks for the value of a quantity. To solve the problems, the model should have a good ability for mathematical reasoning and arithmetic calculations. 

\noindent
\textbf{Task setup.} Also following RAP, we solve the problems by recursively decomposing them into subquestions and then answering them sequentially. After the number of subquestions reaches the limit or the LLM thinks it can answer the whole question, it derives the final answer based on all the subquestions and subanswers. Therefore, we define the states as subanswers, and actions as subquestions. When expanding a state, we ask LLMs to generate a few subquestions as the actions of the current state. The initial value estimation of actions is computed by the normalized probability that LLM answers ``Yes'' compared to answering ``No'' when asked whether this subquestion is useful in the current state.

Like RoT in \textbf{Blocksworld}, we perform MCTS for 20 iterations and BFS with beam size $20$ on 40 samples in the train set, to generate the tree search experiences. We set the threshold $\lambda$ to $0.1$, and still use \textbf{gpt-4} for guideline generation in both RoT and LEAP. The LLMs evaluated in \textbf{GSM8k} are the same as those in \textbf{Blocksworld}: \textbf{phi-2}, \textbf{mistral-7b}, and \textbf{mixtral-8x7b}. 

\begin{table}[t]
    \centering
    \scalebox{0.65}{
    \begin{tabular}{l|cc|c||cc|c||cc|c}
    \toprule
    \multirow{2}[3]{*}{Method} & \multicolumn{3}{c||}{\bf  phi-2} & \multicolumn{3}{c||}{\bf  mistral-7b}  & \multicolumn{3}{c}{ \bf mixtral-8x7b}   \\\rule{0pt}{11px}
    & Base & \bf RoT & LEAP & Base & \bf RoT & LEAP & Base & \bf RoT  & LEAP\\ \midrule
    CoT           & \textbf{48.1} & \textbf{48.1} {\scriptsize $(+0.0\%)$} &  44.4    & 31.2 &  31.8 {\scriptsize \color{red} $(+1.9\%)$} & \textbf{32.4} & 48.7     &  50.2  {\scriptsize \color{red} $(+3.1\%)$} &  \textbf{51.4} \\
    CoT-SC$^{(10)}$           & 66.1 & 64.2 {\scriptsize \color{blue} $(-2.9\%)$}  &  \bf 66.2    & 49.4 &  \textbf{50.3} {\scriptsize \color{red} $(+1.8\%)$}  &  49.5& 69.1      &  71.1  {\scriptsize \color{red} $(+2.9\%)$} &  \textbf{72.7}\\
    BFS$^{(5)}$   & 54.4 &  \textbf{55.8} {\scriptsize \color{red} $(+2.6\%)$} &  51.0 & 46.1 &  \textbf{48.2} {\scriptsize \color{red} $(+4.6\%)$} &  46.4  &  68.0 & 66.7  {\scriptsize \color{blue} $(-2.0\%)$}  &  \bf 67.1   \\
    MCTS$^{(1)}$  & 51.1 &  \textbf{53.5} {\scriptsize \color{red} $(+4.7\%)$} &  50.7 & 42.4 &  \textbf{47.3} {\scriptsize \color{red} $(+11.6\%)$} &  44.8& 65.4     &  \textbf{66.6}  {\scriptsize \color{red} $(+1.8\%)$} &  65.7 \\
    MCTS$^{(10)}$ & 61.9 &  \textbf{62.4} {\scriptsize \color{red} $(+0.8\%)$} &  60.6 & 55.5 &  \textbf{58.9} {\scriptsize \color{red} $(+6.1\%)$} &  56.0 & 77.4     &  \textbf{79.2}  {\scriptsize \color{red} $(+2.3\%)$} & 75.0  \\ \bottomrule
    \end{tabular}
    }
    \caption{{\bf GSM8K} Results.}
    \label{tab:gsm8k-res}
\end{table}

\begin{table*}[t]
    \centering
    \scalebox{0.67}{
    \begin{tabular}{cl|ccc|ccc}
    \toprule
    \multirow{2}{*}{Model}  & \multirow{2}{*}{Method} & \multicolumn{3}{c|}{Encourage Profit} & \multicolumn{3}{c}{Encourage Agree.} \\
      & & Profit$\uparrow$ & Agree.$-$ & Utility$\uparrow$ & Profit$\uparrow$ & Agree.$\uparrow$ & Utility$\uparrow$ \\ \midrule
    \multirowcell{4}{\bf mixtral}  & CoT    & -0.72 & 0.89 & -0.64 &-0.72 & 0.89 & -0.64\\
    & \textbf{RoT}-CoT                      & -0.19 & 1.0 & -0.19 & -0.24 & 1.0 & -0.24 \\ \cdashline{2-8} \rule{0pt}{2.6ex}
    & MCTS$^{(8)}$                          & -0.15 & 1.0 & -0.15 & \bf  0.11 & 1.0 &  \bf 0.11\\
    & \textbf{RoT}-MCTS$^{(8)}$             & \bf 0.03& 1.0 & \bf  0.03 &-0.17 &  1.0 & -0.17\\ \midrule
    \multirowcell{4}{\bf chatgpt} & CoT     & -0.06 & 0.83 & -0.05 & -0.06 & 0.83 & -0.06  \\
    & \textbf{RoT}-CoT                      & \bf  0.80 & 0.72 & 0.58 & 0.43 & 1.0 & 0.43 \\\cdashline{2-8} \rule{0pt}{2.6ex}
     & MCTS$^{(8)}$                          & 0.16 & 0.83 & 0.13 & 0.26 & 1.0 & 0.26\\
    & \textbf{RoT}-MCTS$^{(8)}$             & 0.75 & 0.94 & \bf  0.71 & \bf 0.52 & 1.0 & \bf  0.52 \\ \midrule
    \multirow{4}{*}{\bf gpt-4} & CoT        & -0.46 & 1.0 & -0.46 & -0.46 & 1.0 & -0.46 \\
    & \textbf{RoT}-CoT                      & 0.20 & 0.78 & 0.17 & 0.12 & 0.89 & 0.11 \\\cdashline{2-8} \rule{0pt}{2.6ex}
    & MCTS$^{(8)}$                          & 0.20 & 1.0 & 0.20  & -0.06 & 1.0 & -0.06 \\
    & \textbf{RoT}-MCTS$^{(8)}$             & \bf  0.67 & 0.56 &  \bf 0.38 & \bf 0.59 & 1.0 & \bf 0.59 \\
    \bottomrule
    \end{tabular}
    }
    \caption{Results on \bf CraigslistBargain}
    \label{tab:bargin-res}
\end{table*}
\noindent
\textbf{Results.} As shown in Table \ref{tab:gsm8k-res}, RoT improves the accuracy of various LLMs over baselines. Compared to the non-tree-search-based reflection method, LEAP, RoT can achieve similar performances in CoT and CoT-SC$^{(10)}$, and obtain much better performance in tree-search-based methods.  However, We found that the improvement is less significant than in \textbf{Blocksworld}. One reason is that performing arithmetic calculations accurately is also required to solve the problem, while the guidelines cannot enhance such ability.

\subsection{\textbf{CraigslistBargain}}
\textbf{CraigslistBargain} \citep{bargain} is about bargaining between a buyer and a seller over an item. Each sample consists of the item's description, the buyer's ideal price $p_b$, and the seller's ideal deal price $p_s$. The buyer aims to buy the item at a lower price, while the seller aims at the opposite. However, they may also have a common goal of reaching an agreement. To achieve higher benefits and agreement, a model should be able to compose a good plan to persuade the opponent to adjust their prices. An example is in Figure \ref{fig:bargin-exampl}.

\noindent
\textbf{Task setup.} We define the state as the dialogue history and the action as the concept of the next response to the opponent. When expanding a state, the actions are generated by LLMs. The values of actions are evaluated by roll-out, which requires the current participant to complete the dialogue by both acting as the seller and the buyer. The value of a roll-out is given based on the deal's success and the deal price. Unlike \textbf{Blocksworld} and \textbf{GSM8k}, where the model has full control over transitions between states, the transition in negotiation is also affected by the component. Running a tree search can only get a good action within the current state, instead of a path to the best state. Therefore, the negotiation needs to be finished by multiple rounds of MCTS. Each round involves searching for effective persuasion and transitioning to a new state based on the response from the counterpart.

For RoT, we perform MCTS for 16 iterations on 2 samples to get the experiences and set the threshold $\lambda$ to $0.1$. We use \textbf{gpt-4} to summarize the guidelines. In \textbf{CraigslistBargain}, CoT inquires LLMs to determine an action and generate the response given this action. We evaluate methods as the role of the seller. Since dialogue does not contain a gold answer, we cannot evaluate CoT-SC and LEAP. BFS is also not evaluated due to limited API resources.

We employ \textit{proft} and \textit{utility} to measure how good a successful deal is. Profit from the seller's perspective is defined as $\text{Profit}(p) = 2\cdot\frac{p - (p_s+p_b)}{p_s-p_b},$ where $p$ is the price of a successful deal. This metric provides a linear mapping of the deal price $p$, yielding a value of $-1$ when $p=p_b$ and $1$ when $p=p_s$. In our experimental results, the reported profit is the average profit of successful deals, while the utility is the average profit of all deals, where the profit of an unsuccessful deal is set to 0. Moreover, we use profit in the reward of a dialogue. The reward is defined as 
$l_{penalty}$ if the deal is not successful and Profit$(p)$ if the deal is successful.

In our prior experiments, we observed that without penalizing the seller for unsuccessful deals, it tends to insist on a high price, which leads to high profit but frequent failures in bargains. To address this, we introduce a penalty term, $l_{penalty}$ to control the extent of punishment of an unsuccessful deal. In experiments, we explore two settings, encouraging profit and encouraging agreement, where $l_{penalty}$ is set to $0$ and $1$ respectively. 

Similar to the findings from \cite{yaofu-bargaining}, we observe that smaller LLMs, such as \textbf{phi-2} cannot bargain successfully, as they are unaware of the changing prices. Therefore, we evaluate the framework with larger models including \textbf{mixtral-7x8b}, \textbf{chatgpt} and \textbf{gpt-4} \citep{gpt4}. We use \textbf{gpt-4} as the in all experiments buyer. We evaluate 18 samples when using \textbf{mixtral-8x7b} or \textbf{chatgpt} as the seller and 9 samples when using \textbf{gpt-4} as the seller. 

\noindent 
\textbf{Results.} The results displayed in Table \ref{tab:bargin-res} demonstrate that RoT significantly improves the utilities across most settings. When encouraging the seller to maximize its profit, RoT shows an average absolute profit improvement of $0.55$ while potentially sacrificing some agreement. In settings where agreement is also prioritized, RoT has an average absolute profit improvement of $0.36$ without any compromise in agreement. We surprisingly observe that RoT enables the performance of CoT to match or even surpass that of MCTS. This could be attributed to the guidelines providing models with detailed instructions about responding to the opponent's behavior, reducing the necessity to search for an optimal response. Notably, RoT demonstrates its most impressive performance in \textbf{CraigslistBargin} compared to \textbf{Blocksworld} and \textbf{GSM8k}. One explanation could be that LLMs are less familiar with bargaining, as they are trained not to exhibit overly aggressive behavior, thus making them more inclined to accommodate the opponent's demands. With the guidelines, models can learn to be aggressive and thus highlight the item's value when the opponent wants a lower price instead of directly lowering the price, which leads to higher utility.



\subsection{Search Efficiency Analysis}

\begin{figure}[t]
\centering
\begin{minipage}{0.63\linewidth}
    \centering
    \scalebox{0.67}{
    \begin{tabular}{ll|cccccc}
    \toprule
    Model & Method & Step 2 & Step 4 & Step 6 & Step 8 & Step 10 \\ \midrule
    \multirow{3}{*}{\bf phi-2}  & MCTS &\bf 75.6      & 55.8      & 27.2  & 6.0 & 1.9\\
    & \multirow{2}{*}{RoT-MCTS}                      & 74.7   & \bf 59.1  &\bf 33.1  &\bf 9.7 & \bf 2.0 \\
    & & {\small \color{blue} (-1.1\%)}     & {\small \color{red} (+5.9\%)}      & {\small\color{red} (+21.7\%) } & {\small \color{red} (+61.7\%)} & {\small \color{red} (+5.3\%)} \\  \midrule
    \multirow{3}{*}{\bf mistral-7b}  & MCTS    &\bf 77.7      & 76.5      & 57.7  & 30.4 & 16.2\\
    & \multirow{2}{*}{RoT-MCTS}                           & 76.3   & \bf 79.8  & \bf 58.1  &\bf 37.6 & \bf 22.3 \\
    & & {\small \color{blue} (-1.8\%)}     & {\small \color{red} (+4.3\%)}      &{\small \color{red} (+7.0\%) }  &{\small \color{red} (+23.7\%)}& {\small \color{red} (+37.7\%)}\\ \midrule
    \multirow{3}{*}{\bf mixtral-8x7b}  & MCTS & 82.1& 80.3 & \bf 58.0 &29.5 & 20.3\\
    & \multirow{2}{*}{RoT-MCTS} &\bf 82.7 &\bf 85.1 & 56.7 &\bf 34.4 & \bf 24.1\\
    & & {\small  \color{red} (+0.7\%)}     & {\small \color{red} (+6.0\%)}      &{\small -\color{red} (+3.3\%)}  &{\small \color{red} (+16.6\%)} & {\small \color{red} (+18.7\%)}\\
    \bottomrule
    \end{tabular}
    }
    \captionof{table}{AUC of \textbf{Blocksworld}.}
    \label{tab:bw-auc}
\end{minipage}
\begin{minipage}{0.36\linewidth}
    \centering
    \includegraphics[width=0.9\linewidth, trim=20 20 10 0]{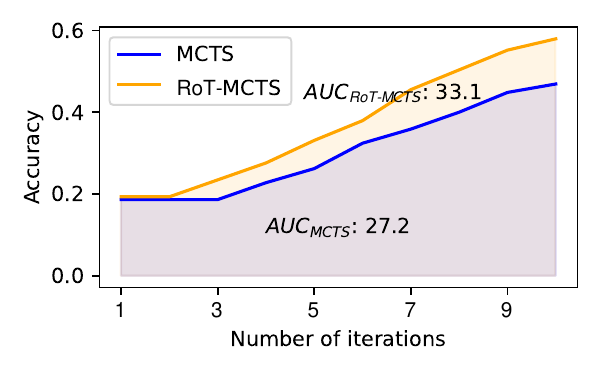}
    \caption{AUC on the step-6 split of \textbf{Blocksworld} using \textbf{phi-2}.}
    \label{fig:auc-sample}
\end{minipage}
\end{figure}

To demonstrate the efficacy of our framework in enhancing search efficiency, we evaluate RoT by area under the iteration-accuracy curve (AUC) of MCTS. As shown in Figure \ref{fig:auc-sample}, we vary the number of MCTS iterations from 1 to 10 and calculate the area within 10 iterations. A higher AUC indicates that the method can achieve higher accuracy within fewer MCTS iterations. We access RoT on \textbf{Blocksworld} and \textbf{GSM8k}, as we can evaluate MCTS with a single round of tree search. As shown in Table \ref{tab:bw-auc}, RoT can improve the value of AUC, which means this framework can achieve a better trade-off between search efficiency and accuracy. Results show that the improvement of RoT is more significant when the task gets harder.

\subsection{Important State Selection}

To prove the efficacy of important state selection, we experiment with the impact of reflection targets on the performance of tree searches in \textbf{GSM8k}. We tried MCTS without guidelines, with guidelines summarized from problem-only samples, all experiences, and random states in the search tree. We also change the threshold $\lambda$ to adjust the quantity and quality of the selected states to inspect its impact on tree search. 
\begin{table}[b]
    \centering
    \scalebox{0.68}{
    \begin{tabular}{l|cc}
    \toprule
    Method & No. States & Accuracy \\ \midrule 
    MCTS$^{(1)}$ & - & 42.4 \\ 
    \emph{+ problem samples} & - & 44.7 \\ 
    \emph{+ all experience} & 538 & 45.0 \\ 
    \emph{+ random states} & 219 & 44.2 \\
    \emph{+ imp. states ($\lambda=-0.001$)} & 340 & 46.2\\
    \emph{+ imp. states ($\lambda=0.5$)} & 53 & 45.6 \\ 
    \emph{+ imp. states ($\lambda=0.1$)} & 242 &\bf 47.3 \\ \bottomrule
    \end{tabular}
    }
    \caption{Number of selected decisions and performance of {\bf mistral-7b} on {\bf GSM8k} when employing different selection mechanism for reflection.}
    \label{tab:promising-state}
\end{table}

As shown in Table \ref{tab:promising-state}, MCTS can benefit from guidelines summarized from the problem samples, since an LLM can give a guideline based on its prior knowledge. When reflecting on the important states, the performance of MCTS can be further improved. We list part of the summarized guidelines in Table \ref{tab:samples}. Through observation, we observe that with important state selection, LLMs can summarize more specific guidelines which improve value estimation and next-state prediction of future searches. 

\section{Related Works}
\subsection{Search-based LLM planning and reasoning}
Planning and reasoning tasks typically require a model to make deliberate decisions while considering future consequences. However, naive input-output-based prompting methods cannot make decisions with consideration about the future, leading to poor performance in complex tasks that require multiple steps of reasoning and planning\citep{blocksworld, math-ng, cbbh}. Given that search methods can effectively consider future outcomes by exploration and backtracking, they have become widely adopted to enhance LLM's planning and reasoning abilities. In an initial effort, self-consistency \citep{self-consistency} conducts parallel searches across multiple chains of thoughts and conveys a majority vote to determine the final result. However, since each search process is done individually, the same reasoning steps may be explored multiple times in the search process, resulting in inefficiency. As tree searches can effectively avoid revisiting the sample reasoning process, various tree-search-based prompting methods are proposed to further enhance the model's reasoning and planning ability. For example, tree-of-thoughts \citep{tot} applies breadth-first search and depth-first search in game planning and mathematical reasoning. Other methods employ Monte Carlo Tree Search (MCTS) to enhance reasoning and planning in various tasks including code generation \citep{code-mcts-plan, mcts-code}, dialogue policy planning \citep{mcts-dialogue}, game planning \citep{tot}, and mathematical reasoning \citep{tot, rap}. \cite{astartool} propose to use A* in tool agent and mathematical reasoning. These methods usually rely on LLMs to generate the action, estimate the values of actions, and predict the next state given an action, which controls the exploration process of tree search. However, by investigating the above tree-search-based prompting methods, \cite{treedisc} discovers that the accuracy of value estimation is important to the tree search performance while LLM-based value estimation is unreliable. In RoT, the reflected guidelines can make the value estimation more accurate to achieve better tree searches.
\subsection{Reflection}
Reflection, as seen in methods such as Self-Refine \citep{self-refine} is a method for enhancing models' ability at inference time by refining responses based on feedback. Previous methods primarily focus on refining responses based on a single feedback \citep{self-refine, self-debug} or contrast between multiple models \citep{debate-1, self-contrast}. Although effective, these methods typically concentrate on refining the current sample while discarding the historical reflections, resulting in the loss of past experiences. Consequently, the model cannot form a comprehensive understanding of this task. To address this limitation, various studies propose methods to enhance models' ability by refining policy or prompt based on task-level reflections. ProTeGi \citep{protegi} and PromptAgent \citep{promptagent} optimize the prompt by search algorithms. TRAN \citep{tran}, Grimoire \citep{grimoire}, and LEAP \citep{leap-icl} use guidelines summarized from good or bad generation results to guide the future generation. In our proposed method, we enhance the models' ability by refining the LLM-assisted tree search process with reflection on tree search experiences across various samples. 

\section{Conclusion}
We propose a framework RoT, which prevents the model from making mistakes repeatedly by reflecting on the previous tree search experiences. RoT improves the accuracy and efficiency of LLM tree search by a guideline reflected from previous search experiences. We show that RoT is also beneficial to non-tree-search-based prompting methods including BFS and MCTS by providing models with better knowledge prior. Experiments demonstrate that RoT can significantly enhance the performance of LLM prompting methods, especially for tree-search-based prompting methods.

\bibliography{colm2024_conference}

\begin{thebibliography}{33}
\providecommand{\natexlab}[1]{#1}
\providecommand{\url}[1]{\texttt{#1}}
\expandafter\ifx\csname urlstyle\endcsname\relax
  \providecommand{\doi}[1]{doi: #1}\else
  \providecommand{\doi}{doi: \begingroup \urlstyle{rm}\Url}\fi

\bibitem[Anthony et~al.(2017)Anthony, Tian, and Barber]{expertiter}
Thomas Anthony, Zheng Tian, and David Barber.
\newblock Thinking fast and slow with deep learning and tree search.
\newblock \emph{Advances in neural information processing systems}, 30, 2017.

\bibitem[Chen et~al.(2024{\natexlab{a}})Chen, Song, Yu, Li, Wang, Xiong, and Tang]{grimoire}
Ding Chen, Shichao Song, Qingchen Yu, Zhiyu Li, Wenjin Wang, Feiyu Xiong, and Bo~Tang.
\newblock Grimoire is all you need for enhancing large language models.
\newblock \emph{CoRR}, abs/2401.03385, 2024{\natexlab{a}}.
\newblock \doi{10.48550/ARXIV.2401.03385}.
\newblock URL \url{https://doi.org/10.48550/arXiv.2401.03385}.

\bibitem[Chen et~al.(2023)Chen, Lin, Sch{\"{a}}rli, and Zhou]{self-debug}
Xinyun Chen, Maxwell Lin, Nathanael Sch{\"{a}}rli, and Denny Zhou.
\newblock Teaching large language models to self-debug.
\newblock \emph{CoRR}, abs/2304.05128, 2023.
\newblock \doi{10.48550/ARXIV.2304.05128}.
\newblock URL \url{https://doi.org/10.48550/arXiv.2304.05128}.

\bibitem[Chen et~al.(2024{\natexlab{b}})Chen, White, Mooney, Payani, Su, and Sun]{treedisc}
Ziru Chen, Michael White, Raymond Mooney, Ali Payani, Yu~Su, and Huan Sun.
\newblock When is tree search useful for llm planning? it depends on the discriminator.
\newblock 2024{\natexlab{b}}.
\newblock URL \url{https://api.semanticscholar.org/CorpusID:267740392}.

\bibitem[Cobbe et~al.(2021)Cobbe, Kosaraju, Bavarian, Chen, Jun, Kaiser, Plappert, Tworek, Hilton, Nakano, Hesse, and Schulman]{gsm8k}
Karl Cobbe, Vineet Kosaraju, Mohammad Bavarian, Mark Chen, Heewoo Jun, Lukasz Kaiser, Matthias Plappert, Jerry Tworek, Jacob Hilton, Reiichiro Nakano, Christopher Hesse, and John Schulman.
\newblock Training verifiers to solve math word problems.
\newblock \emph{CoRR}, abs/2110.14168, 2021.
\newblock URL \url{https://arxiv.org/abs/2110.14168}.

\bibitem[Coulom(2006)]{mcts2}
R{\'{e}}mi Coulom.
\newblock Efficient selectivity and backup operators in monte-carlo tree search.
\newblock In H.~Jaap van~den Herik, Paolo Ciancarini, and H.~H. L.~M. Donkers (eds.), \emph{Computers and Games, 5th International Conference, {CG} 2006, Turin, Italy, May 29-31, 2006. Revised Papers}, volume 4630 of \emph{Lecture Notes in Computer Science}, pp.\  72--83. Springer, 2006.
\newblock \doi{10.1007/978-3-540-75538-8\_7}.
\newblock URL \url{https://doi.org/10.1007/978-3-540-75538-8\_7}.

\bibitem[Du et~al.(2023)Du, Li, Torralba, Tenenbaum, and Mordatch]{debate-1}
Yilun Du, Shuang Li, Antonio Torralba, Joshua~B. Tenenbaum, and Igor Mordatch.
\newblock Improving factuality and reasoning in language models through multiagent debate.
\newblock \emph{CoRR}, abs/2305.14325, 2023.
\newblock \doi{10.48550/ARXIV.2305.14325}.
\newblock URL \url{https://doi.org/10.48550/arXiv.2305.14325}.

\bibitem[Fu et~al.(2023)Fu, Peng, Khot, and Lapata]{yaofu-bargaining}
Yao Fu, Hao Peng, Tushar Khot, and Mirella Lapata.
\newblock Improving language model negotiation with self-play and in-context learning from {AI} feedback.
\newblock \emph{CoRR}, abs/2305.10142, 2023.
\newblock \doi{10.48550/ARXIV.2305.10142}.
\newblock URL \url{https://doi.org/10.48550/arXiv.2305.10142}.

\bibitem[Hao et~al.(2023)Hao, Gu, Ma, Hong, Wang, Wang, and Hu]{rap}
Shibo Hao, Yi~Gu, Haodi Ma, Joshua~Jiahua Hong, Zhen Wang, Daisy~Zhe Wang, and Zhiting Hu.
\newblock Reasoning with language model is planning with world model.
\newblock In Houda Bouamor, Juan Pino, and Kalika Bali (eds.), \emph{Proceedings of the 2023 Conference on Empirical Methods in Natural Language Processing, {EMNLP} 2023, Singapore, December 6-10, 2023}, pp.\  8154--8173. Association for Computational Linguistics, 2023.
\newblock URL \url{https://aclanthology.org/2023.emnlp-main.507}.

\bibitem[Hart et~al.(1968)Hart, Nilsson, and Raphael]{astar}
Peter~E. Hart, Nils~J. Nilsson, and Bertram Raphael.
\newblock A formal basis for the heuristic determination of minimum cost paths.
\newblock \emph{{IEEE} Trans. Syst. Sci. Cybern.}, 4\penalty0 (2):\penalty0 100--107, 1968.
\newblock \doi{10.1109/TSSC.1968.300136}.
\newblock URL \url{https://doi.org/10.1109/TSSC.1968.300136}.

\bibitem[He et~al.(2018)He, Chen, Balakrishnan, and Liang]{bargain}
He~He, Derek Chen, Anusha Balakrishnan, and Percy Liang.
\newblock Decoupling strategy and generation in negotiation dialogues.
\newblock In Ellen Riloff, David Chiang, Julia Hockenmaier, and Jun'ichi Tsujii (eds.), \emph{Proceedings of the 2018 Conference on Empirical Methods in Natural Language Processing, Brussels, Belgium, October 31 - November 4, 2018}, pp.\  2333--2343. Association for Computational Linguistics, 2018.
\newblock \doi{10.18653/V1/D18-1256}.
\newblock URL \url{https://doi.org/10.18653/v1/d18-1256}.

\bibitem[Jiang et~al.(2023)Jiang, Sablayrolles, Mensch, Bamford, Chaplot, de~Las~Casas, Bressand, Lengyel, Lample, Saulnier, Lavaud, Lachaux, Stock, Scao, Lavril, Wang, Lacroix, and Sayed]{mistral}
Albert~Q. Jiang, Alexandre Sablayrolles, Arthur Mensch, Chris Bamford, Devendra~Singh Chaplot, Diego de~Las~Casas, Florian Bressand, Gianna Lengyel, Guillaume Lample, Lucile Saulnier, L{\'{e}}lio~Renard Lavaud, Marie{-}Anne Lachaux, Pierre Stock, Teven~Le Scao, Thibaut Lavril, Thomas Wang, Timoth{\'{e}}e Lacroix, and William~El Sayed.
\newblock Mistral 7b.
\newblock \emph{CoRR}, abs/2310.06825, 2023.
\newblock \doi{10.48550/ARXIV.2310.06825}.
\newblock URL \url{https://doi.org/10.48550/arXiv.2310.06825}.

\bibitem[Jiang et~al.(2024)Jiang, Sablayrolles, Roux, Mensch, Savary, Bamford, Chaplot, de~Las~Casas, Hanna, Bressand, Lengyel, Bour, Lample, Lavaud, Saulnier, Lachaux, Stock, Subramanian, Yang, Antoniak, Scao, Gervet, Lavril, Wang, Lacroix, and Sayed]{mixtral}
Albert~Q. Jiang, Alexandre Sablayrolles, Antoine Roux, Arthur Mensch, Blanche Savary, Chris Bamford, Devendra~Singh Chaplot, Diego de~Las~Casas, Emma~Bou Hanna, Florian Bressand, Gianna Lengyel, Guillaume Bour, Guillaume Lample, L{\'{e}}lio~Renard Lavaud, Lucile Saulnier, Marie{-}Anne Lachaux, Pierre Stock, Sandeep Subramanian, Sophia Yang, Szymon Antoniak, Teven~Le Scao, Th{\'{e}}ophile Gervet, Thibaut Lavril, Thomas Wang, Timoth{\'{e}}e Lacroix, and William~El Sayed.
\newblock Mixtral of experts.
\newblock \emph{CoRR}, abs/2401.04088, 2024.
\newblock \doi{10.48550/ARXIV.2401.04088}.
\newblock URL \url{https://doi.org/10.48550/arXiv.2401.04088}.

\bibitem[Kocsis \& Szepesv{\'{a}}ri(2006)Kocsis and Szepesv{\'{a}}ri]{mcts1}
Levente Kocsis and Csaba Szepesv{\'{a}}ri.
\newblock Bandit based monte-carlo planning.
\newblock In Johannes F{\"{u}}rnkranz, Tobias Scheffer, and Myra Spiliopoulou (eds.), \emph{Machine Learning: {ECML} 2006, 17th European Conference on Machine Learning, Berlin, Germany, September 18-22, 2006, Proceedings}, volume 4212 of \emph{Lecture Notes in Computer Science}, pp.\  282--293. Springer, 2006.
\newblock \doi{10.1007/11871842\_29}.
\newblock URL \url{https://doi.org/10.1007/11871842\_29}.

\bibitem[Kojima et~al.(2022)Kojima, Gu, Reid, Matsuo, and Iwasawa]{cot2}
Takeshi Kojima, Shixiang~Shane Gu, Machel Reid, Yutaka Matsuo, and Yusuke Iwasawa.
\newblock Large language models are zero-shot reasoners.
\newblock In Sanmi Koyejo, S.~Mohamed, A.~Agarwal, Danielle Belgrave, K.~Cho, and A.~Oh (eds.), \emph{Advances in Neural Information Processing Systems 35: Annual Conference on Neural Information Processing Systems 2022, NeurIPS 2022, New Orleans, LA, USA, November 28 - December 9, 2022}, 2022.
\newblock URL \url{http://papers.nips.cc/paper\_files/paper/2022/hash/8bb0d291acd4acf06ef112099c16f326-Abstract-Conference.html}.

\bibitem[Li et~al.(2023{\natexlab{a}})Li, Lin, Zhang, Fu, Chen, Lou, and Chen]{math-ng}
Yifei Li, Zeqi Lin, Shizhuo Zhang, Qiang Fu, Bei Chen, Jian{-}Guang Lou, and Weizhu Chen.
\newblock Making language models better reasoners with step-aware verifier.
\newblock In Anna Rogers, Jordan~L. Boyd{-}Graber, and Naoaki Okazaki (eds.), \emph{Proceedings of the 61st Annual Meeting of the Association for Computational Linguistics (Volume 1: Long Papers), {ACL} 2023, Toronto, Canada, July 9-14, 2023}, pp.\  5315--5333. Association for Computational Linguistics, 2023{\natexlab{a}}.
\newblock \doi{10.18653/V1/2023.ACL-LONG.291}.
\newblock URL \url{https://doi.org/10.18653/v1/2023.acl-long.291}.

\bibitem[Li et~al.(2023{\natexlab{b}})Li, Bubeck, Eldan, Del~Giorno, Gunasekar, and Lee]{phi-2}
Yuanzhi Li, S{\'e}bastien Bubeck, Ronen Eldan, Allie Del~Giorno, Suriya Gunasekar, and Yin~Tat Lee.
\newblock Textbooks are all you need ii: \textbf{phi-1.5} technical report.
\newblock \emph{arXiv preprint arXiv:2309.05463}, 2023{\natexlab{b}}.

\bibitem[Madaan et~al.(2023)Madaan, Tandon, Gupta, Hallinan, Gao, Wiegreffe, Alon, Dziri, Prabhumoye, Yang, Welleck, Majumder, Gupta, Yazdanbakhsh, and Clark]{self-refine}
Aman Madaan, Niket Tandon, Prakhar Gupta, Skyler Hallinan, Luyu Gao, Sarah Wiegreffe, Uri Alon, Nouha Dziri, Shrimai Prabhumoye, Yiming Yang, Sean Welleck, Bodhisattwa~Prasad Majumder, Shashank Gupta, Amir Yazdanbakhsh, and Peter Clark.
\newblock Self-refine: Iterative refinement with self-feedback.
\newblock \emph{CoRR}, abs/2303.17651, 2023.
\newblock \doi{10.48550/ARXIV.2303.17651}.
\newblock URL \url{https://doi.org/10.48550/arXiv.2303.17651}.

\bibitem[OpenAI(2023)]{gpt4}
OpenAI.
\newblock {GPT-4} technical report.
\newblock \emph{CoRR}, abs/2303.08774, 2023.
\newblock \doi{10.48550/ARXIV.2303.08774}.
\newblock URL \url{https://doi.org/10.48550/arXiv.2303.08774}.

\bibitem[Pryzant et~al.(2023)Pryzant, Iter, Li, Lee, Zhu, and Zeng]{protegi}
Reid Pryzant, Dan Iter, Jerry Li, Yin~Tat Lee, Chenguang Zhu, and Michael Zeng.
\newblock Automatic prompt optimization with "gradient descent" and beam search.
\newblock In Houda Bouamor, Juan Pino, and Kalika Bali (eds.), \emph{Proceedings of the 2023 Conference on Empirical Methods in Natural Language Processing, {EMNLP} 2023, Singapore, December 6-10, 2023}, pp.\  7957--7968. Association for Computational Linguistics, 2023.
\newblock URL \url{https://aclanthology.org/2023.emnlp-main.494}.

\bibitem[Suzgun et~al.(2023)Suzgun, Scales, Sch{\"{a}}rli, Gehrmann, Tay, Chung, Chowdhery, Le, Chi, Zhou, and Wei]{cbbh}
Mirac Suzgun, Nathan Scales, Nathanael Sch{\"{a}}rli, Sebastian Gehrmann, Yi~Tay, Hyung~Won Chung, Aakanksha Chowdhery, Quoc~V. Le, Ed~H. Chi, Denny Zhou, and Jason Wei.
\newblock Challenging big-bench tasks and whether chain-of-thought can solve them.
\newblock In Anna Rogers, Jordan~L. Boyd{-}Graber, and Naoaki Okazaki (eds.), \emph{Findings of the Association for Computational Linguistics: {ACL} 2023, Toronto, Canada, July 9-14, 2023}, pp.\  13003--13051. Association for Computational Linguistics, 2023.
\newblock \doi{10.18653/V1/2023.FINDINGS-ACL.824}.
\newblock URL \url{https://doi.org/10.18653/v1/2023.findings-acl.824}.

\bibitem[Valmeekam et~al.(2023)Valmeekam, Marquez, Sreedharan, and Kambhampati]{blocksworld}
Karthik Valmeekam, Matthew Marquez, Sarath Sreedharan, and Subbarao Kambhampati.
\newblock On the planning abilities of large language models - a critical investigation.
\newblock In \emph{Thirty-seventh Conference on Neural Information Processing Systems}, 2023.
\newblock URL \url{https://openreview.net/forum?id=X6dEqXIsEW}.

\bibitem[Wang et~al.(2023{\natexlab{a}})Wang, Li, Wang, Bai, Luo, Zhang, Jojic, Xing, and Hu]{promptagent}
Xinyuan Wang, Chenxi Li, Zhen Wang, Fan Bai, Haotian Luo, Jiayou Zhang, Nebojsa Jojic, Eric~P. Xing, and Zhiting Hu.
\newblock Promptagent: Strategic planning with language models enables expert-level prompt optimization.
\newblock \emph{CoRR}, abs/2310.16427, 2023{\natexlab{a}}.
\newblock \doi{10.48550/ARXIV.2310.16427}.
\newblock URL \url{https://doi.org/10.48550/arXiv.2310.16427}.

\bibitem[Wang et~al.(2023{\natexlab{b}})Wang, Wei, Schuurmans, Le, Chi, Narang, Chowdhery, and Zhou]{self-consistency}
Xuezhi Wang, Jason Wei, Dale Schuurmans, Quoc~V. Le, Ed~H. Chi, Sharan Narang, Aakanksha Chowdhery, and Denny Zhou.
\newblock Self-consistency improves chain of thought reasoning in language models.
\newblock In \emph{The Eleventh International Conference on Learning Representations, {ICLR} 2023, Kigali, Rwanda, May 1-5, 2023}. OpenReview.net, 2023{\natexlab{b}}.
\newblock URL \url{https://openreview.net/pdf?id=1PL1NIMMrw}.

\bibitem[Wei et~al.(2022)Wei, Wang, Schuurmans, Bosma, Ichter, Xia, Chi, Le, and Zhou]{cot}
Jason Wei, Xuezhi Wang, Dale Schuurmans, Maarten Bosma, Brian Ichter, Fei Xia, Ed~H. Chi, Quoc~V. Le, and Denny Zhou.
\newblock Chain-of-thought prompting elicits reasoning in large language models.
\newblock In Sanmi Koyejo, S.~Mohamed, A.~Agarwal, Danielle Belgrave, K.~Cho, and A.~Oh (eds.), \emph{Advances in Neural Information Processing Systems 35: Annual Conference on Neural Information Processing Systems 2022, NeurIPS 2022, New Orleans, LA, USA, November 28 - December 9, 2022}, 2022.
\newblock URL \url{http://papers.nips.cc/paper\_files/paper/2022/hash/9d5609613524ecf4f15af0f7b31abca4-Abstract-Conference.html}.

\bibitem[Yang et~al.(2023{\natexlab{a}})Yang, Prabhakar, Narasimhan, and Yao]{mcts-code}
John Yang, Akshara Prabhakar, Karthik~R Narasimhan, and Shunyu Yao.
\newblock Intercode: Standardizing and benchmarking interactive coding with execution feedback.
\newblock In \emph{Thirty-seventh Conference on Neural Information Processing Systems Datasets and Benchmarks Track}, 2023{\natexlab{a}}.
\newblock URL \url{https://openreview.net/forum?id=fvKaLF1ns8}.

\bibitem[Yang et~al.(2023{\natexlab{b}})Yang, Li, and Liu]{tran}
Zeyuan Yang, Peng Li, and Yang Liu.
\newblock Failures pave the way: Enhancing large language models through tuning-free rule accumulation.
\newblock In Houda Bouamor, Juan Pino, and Kalika Bali (eds.), \emph{Proceedings of the 2023 Conference on Empirical Methods in Natural Language Processing, {EMNLP} 2023, Singapore, December 6-10, 2023}, pp.\  1751--1777. Association for Computational Linguistics, 2023{\natexlab{b}}.
\newblock URL \url{https://aclanthology.org/2023.emnlp-main.109}.

\bibitem[Yao et~al.(2023)Yao, Yu, Zhao, Shafran, Griffiths, Cao, and Narasimhan]{tot}
Shunyu Yao, Dian Yu, Jeffrey Zhao, Izhak Shafran, Thomas~L. Griffiths, Yuan Cao, and Karthik Narasimhan.
\newblock Tree of thoughts: Deliberate problem solving with large language models.
\newblock \emph{CoRR}, abs/2305.10601, 2023.
\newblock \doi{10.48550/ARXIV.2305.10601}.
\newblock URL \url{https://doi.org/10.48550/arXiv.2305.10601}.

\bibitem[Yu et~al.(2023)Yu, Chen, and Yu]{mcts-dialogue}
Xiao Yu, Maximillian Chen, and Zhou Yu.
\newblock Prompt-based monte-carlo tree search for goal-oriented dialogue policy planning.
\newblock In Houda Bouamor, Juan Pino, and Kalika Bali (eds.), \emph{Proceedings of the 2023 Conference on Empirical Methods in Natural Language Processing, {EMNLP} 2023, Singapore, December 6-10, 2023}, pp.\  7101--7125. Association for Computational Linguistics, 2023.
\newblock URL \url{https://aclanthology.org/2023.emnlp-main.439}.

\bibitem[Zhang et~al.(2023)Zhang, Chen, Shen, Ding, Tenenbaum, and Gan]{code-mcts-plan}
Shun Zhang, Zhenfang Chen, Yikang Shen, Mingyu Ding, Joshua~B. Tenenbaum, and Chuang Gan.
\newblock Planning with large language models for code generation.
\newblock In \emph{The Eleventh International Conference on Learning Representations, {ICLR} 2023, Kigali, Rwanda, May 1-5, 2023}. OpenReview.net, 2023.
\newblock URL \url{https://openreview.net/pdf?id=Lr8cOOtYbfL}.

\bibitem[Zhang et~al.(2024{\natexlab{a}})Zhang, Madaan, Gao, Zheng, Mishra, Yang, Tandon, and Alon]{leap-icl}
Tianjun Zhang, Aman Madaan, Luyu Gao, Steven Zheng, Swaroop Mishra, Yiming Yang, Niket Tandon, and Uri Alon.
\newblock In-context principle learning from mistakes, 2024{\natexlab{a}}.

\bibitem[Zhang et~al.(2024{\natexlab{b}})Zhang, Shen, Wu, Peng, Wang, Zhuang, and Lu]{self-contrast}
Wenqi Zhang, Yongliang Shen, Linjuan Wu, Qiuying Peng, Jun Wang, Yueting Zhuang, and Weiming Lu.
\newblock Self-contrast: Better reflection through inconsistent solving perspectives.
\newblock \emph{CoRR}, abs/2401.02009, 2024{\natexlab{b}}.
\newblock \doi{10.48550/ARXIV.2401.02009}.
\newblock URL \url{https://doi.org/10.48550/arXiv.2401.02009}.

\bibitem[Zhuang et~al.(2024)Zhuang, Chen, Yu, Mitra, Bursztyn, Rossi, Sarkhel, and Zhang]{astartool}
Yuchen Zhuang, Xiang Chen, Tong Yu, Saayan Mitra, Victor Bursztyn, Ryan~A. Rossi, Somdeb Sarkhel, and Chao Zhang.
\newblock Toolchain*: Efficient action space navigation in large language models with a* search.
\newblock In \emph{The Twelfth International Conference on Learning Representations}, 2024.
\newblock URL \url{https://openreview.net/forum?id=B6pQxqUcT8}.

\end{thebibliography}
\bibliographystyle{colm2024_conference}

\appendix
\section{Limitations}
RoT has several limitations. It requires models to have good overall capabilities. A strong model is required to summarize a meaningful guideline. Furthermore, the model used to perform tree search should have good instruction-following ability. Otherwise, it may not follow the guidelines and thus the guidelines will be useless. To select important states and reflect meaningful guidelines, RoT requires accurate value estimation of states in trees. This requires a strong value estimation method or in MCTS, a large number of iterations.

\section{Additional Experiments Details}
\label{sec:appendix}

\subsection{LLM Input \& Output Examples} \label{sec:block-sample}
We show the input \& output examples in \textbf{Blocksworld}. For prompts in \textbf{GSM8k} and \textbf{CraigslistBargain}, see our GitHub repository. We use prompts from RAP \citep{rap} in \textbf{blocksworld}.

\noindent\textbf{Action Evaluation}
\begin{Verbatim}[commandchars=\\\{\},breaklines=true,breaksymbolleft=]
Input:
I am playing with a set of blocks where I need to arrange the blocks into stacks. Here are the actions I can do

Pick up a block
Unstack a block from on top of another block
Put down a block
Stack a block on top of another block

I have the following restrictions on my actions:
I can only pick up or unstack one block at a time.
I can only pick up or unstack a block if my hand is empty.
...

\textcolor{red}{To effectively manage the BlocksWorld game environment and achieve higher rewards, the following consolidated policy guideline should be applied by the agent:\\1. The agent should focus on unstacking or picking up blocks that are essential for constructing the goal configuration, especially if they are not already positioned correctly. It should endeavor to clear blocks that are part of the goal configuration by ensuring that no other blocks are stacked on top of them and are thus free to be manipulated.\\2. When the agent's hand is empty, it should prioritize picking up blocks that will contribute to the immediate construction of a correct sub-configuration which leads towards the overall goal configuration. Preferentially, blocks should be picked up in an order that reflects their final position in the goal state to minimize additional moves.\\3. If the agent is holding a block, it must assess whether placing it immediately contributes to the goal configuration. If the action is valid and moves closer to achieving the goal, it should place the block accordingly. Otherwise, the agent should put it down on the table in a clear space to keep options open for subsequent actions.\\...}

Please evaluate whether the given action is a good one under certain conditions.

[STATEMENT]
As initial conditions I have that, the red block is clear, the yellow block is clear, the hand is empty, the red block is on top of the blue block, the yellow block is on top of the orange block, the blue block is on the table and the orange block is on the table.
My goal is to have that the orange block is on top of the red block.
[ACTION]
unstack the red block from on top of the blue block
[EVALUATION]
bad
...
[STATEMENT]
\textcolor{blue}{As initial conditions I have that, the red block is clear, the yellow block is clear, the hand is empty, the red block is on top of the blue block, the yellow block is on top of the orange block, the blue block is on the table and the orange block is on the table.}
\textcolor{blue}{My goal is to have that the red block is on top of the orange block and the orange block is on top of the blue block.}
[ACTION]
\textcolor{green}{unstack the yellow block from on top of the orange block}
[EVALUATION]

Outputs:
Normalized probability of outputting good vs bad.

\end{Verbatim}
The red part is the guideline, the blue part is the description of the current state, and the green part is the action to evaluate.
\newpage
\subsection{Full Blocksworld Results}
\begin{table}[h]

    \centering
    \scalebox{0.5}{
    \begin{tabular}{ll|ccccccccccccccc}
    \toprule
     & \multirow{2}[3]{*}{Method} & \multicolumn{3}{c}{Step 2} & \multicolumn{3}{c}{Step 4}  & \multicolumn{3}{c}{Step 6}  & \multicolumn{3}{c}{Step 8}  & \multicolumn{3}{c}{Step 10}  \\ \cmidrule(lr){3-5} \cmidrule(lr){6-8}\cmidrule(lr){9-11}\cmidrule(lr){12-14}\cmidrule(lr){15-17}
    & & Base & \bf  RoT & \textit{LEAP}&  Base & \bf  RoT & \textit{LEAP} &  Base & \bf  RoT & \textit{LEAP} &  Base & \bf  RoT & \textit{LEAP} &  Base & \bf  RoT & \textit{LEAP} \\ \midrule
    \multirow{5}{*}{\rotatebox{90}{\textbf{phi-2\ \ \ \ }}} & CoT & 8.1 & {13.5} {\scriptsize {\color{red}(+66.7\%)}} & \small\textit{15.4} & 1.3 & {3.9} {\scriptsize {\color{red}(+200.0\%)}} & \small\textit{1.7} & 0.0 & 0.0 {\scriptsize {\color{black}(+0.0\%)}} & \small\textit{0.0} & 0.0 & 0.0 {\scriptsize {\color{black}(+0.0\%)}} & \small\textit{0.0} & 0.0 & 0.0 {\scriptsize {\color{black}(+0.0\%)}} & \small\textit{0.0}\\
 & CoT-Pass@10 & 43.2 & {54.1} {\scriptsize {\color{red}(+25.2\%)}} & \small\textit{56.8} & 5.3 & {15.8} {\scriptsize {\color{red}(+198.1\%)}} & \small\textit{13.2} & 1.4 & 1.4 {\scriptsize {\color{black}(+0.0\%)}} & \small\textit{2.1} & 0.0 & {0.7} {\scriptsize {\color{black}(+0.0\%)}} & \small\textit{0.0} & 0.0 & 0.0 {\scriptsize {\color{black}(+0.0\%)}} & \small\textit{0.0}\\
 & BFS$^{(5)}$ & 97.3 & {100.0} {\scriptsize {\color{red}(+2.8\%)}} & \small\textit{100.0} & 61.8 & {69.7} {\scriptsize {\color{red}(+12.8\%)}} & \small\textit{67.1} & 25.5 & {29.0} {\scriptsize {\color{red}(+13.7\%)}} & \small\textit{33.1} & 11.2 & {16.1} {\scriptsize {\color{red}(+43.8\%)}} & \small\textit{14.0} & 5.8 & {8.7} {\scriptsize {\color{red}(+50.0\%)}} & \small\textit{5.8}\\
 & MCTS$^{(1)}$ & 55.6 & {59.5} {\scriptsize {\color{red}(+7.0\%)}} & \small\textit{59.5} & 30.3 & {36.8} {\scriptsize {\color{red}(+21.5\%)}} & \small\textit{28.9} & 17.9 & {18.6} {\scriptsize {\color{red}(+3.9\%)}} & \small\textit{22.8} & 4.2 & {6.3} {\scriptsize {\color{red}(+50.0\%)}} & \small\textit{4.2} & 1.9 & 1.9 {\scriptsize {\color{black}(+0.0\%)}} & \small\textit{1.9}\\
 & MCTS$^{(10)}$ & {89.2} & 86.5 {\scriptsize {\color{blue}(-3.0\%)}} & \small\textit{89.2} & 84.2 & {85.5} {\scriptsize {\color{red}(+1.5\%)}} & \small\textit{80.3} & 46.9 & {55.2} {\scriptsize {\color{red}(+17.7\%)}} & \small\textit{53.1} & 11.2 & {17.5} {\scriptsize {\color{red}(+56.2\%)}} & \small\textit{14.7} & 2.9 & {3.9} {\scriptsize {\color{red}(+34.5\%)}} & \small\textit{2.9}\\
\midrule
\multirow{5}{*}{\rotatebox{90}{\textbf{mistral-7b\ \ \ }}} & CoT & 35.1 & {43.2} {\scriptsize {\color{red}(+23.1\%)}} & \small\textit{27.0} & 5.3 & {10.5} {\scriptsize {\color{red}(+98.1\%)}} & \small\textit{10.9} & 3.4 & {6.2} {\scriptsize {\color{red}(+82.4\%)}} & \small\textit{3.8} & 0.0 & 0.0 {\scriptsize {\color{black}(+0.0\%)}} & \small\textit{0.0} & 0.0 & {1.0} {\scriptsize {\color{black}(+0.0\%)}} & \small\textit{1.2}\\
 & CoT-Pass@10 & {83.8} & 81.1 {\scriptsize {\color{blue}(-3.2\%)}} & \small\textit{81.1} & 40.8 & {48.7} {\scriptsize {\color{red}(+19.4\%)}} & \small\textit{55.3} & {24.1} & 21.4 {\scriptsize {\color{blue}(-11.2\%)}} & \small\textit{26.9} & 4.9 & {10.5} {\scriptsize {\color{red}(+114.3\%)}} & \small\textit{9.0} & 7.8 & {11.7} {\scriptsize {\color{red}(+50.0\%)}} & \small\textit{10.7}\\
 & BFS$^{(5)}$ & 100.0 & 100.0 {\scriptsize {\color{black}(+0.0\%)}} & \small\textit{100.0} & 86.8 & {89.5} {\scriptsize {\color{red}(+3.1\%)}} & \small\textit{89.5} & {64.1} & 61.4 {\scriptsize {\color{blue}(-4.2\%)}} & \small\textit{62.1} & 42.0 & {42.7} {\scriptsize {\color{red}(+1.7\%)}} & \small\textit{42.0} & 32.0 & {33.0} {\scriptsize {\color{red}(+3.1\%)}} & \small\textit{33.0}\\
 & MCTS$^{(1)}$ & 64.9 & {67.6} {\scriptsize {\color{red}(+4.2\%)}} & \small\textit{75.7} & 59.2 & {61.8} {\scriptsize {\color{red}(+4.4\%)}} & \small\textit{61.8} & 40.0 & {43.4} {\scriptsize {\color{red}(+8.5\%)}} & \small\textit{41.4} & 14.5 & {24.5} {\scriptsize {\color{red}(+69.0\%)}} & \small\textit{21.0} & 12.6 & {21.4} {\scriptsize {\color{red}(+69.8\%)}} & \small\textit{18.4}\\
 & MCTS$^{(10)}$ & 78.4 & {83.8} {\scriptsize {\color{red}(+6.9\%)}} & \small\textit{89.2} & 89.5 & {92.1} {\scriptsize {\color{red}(+2.9\%)}} & \small\textit{94.7} & {76.6} & 76.0 {\scriptsize {\color{blue}(-0.8\%)}} & \small\textit{77.2} & 55.2 & {59.4} {\scriptsize {\color{red}(+7.6\%)}} & \small\textit{58.0} & 29.1 & {34.0} {\scriptsize {\color{red}(+16.8\%)}} & \small\textit{30.1}\\
\midrule
\multirow{5}{*}{\rotatebox{90}{\textbf{mixtral-8x7b\ }}} & CoT & 29.7 & {45.9} {\scriptsize {\color{red}(+54.5\%)}} & \small\textit{34.3} & 9.2 & {18.4} {\scriptsize {\color{red}(+100.0\%)}} & \small\textit{13.0} & 11.0 & {11.7} {\scriptsize {\color{red}(+6.4\%)}} & \small\textit{9.4} & 2.1 & {4.9} {\scriptsize {\color{red}(+133.3\%)}} & \small\textit{2.8} & 2.9 & {3.9} {\scriptsize {\color{red}(+34.5\%)}} & \small\textit{3.3}\\
 & CoT-Pass@10 & 81.1 & {86.5} {\scriptsize {\color{red}(+6.7\%)}} & \small\textit{89.2} & 63.2 & {71.1} {\scriptsize {\color{red}(+12.5\%)}} & \small\textit{61.8} & 54.5 & 54.5 {\scriptsize {\color{black}(+0.0\%)}} & \small\textit{53.1} & 17.5 & {29.4} {\scriptsize {\color{red}(+68.0\%)}} & \small\textit{21.7} & 20.4 & {27.2} {\scriptsize {\color{red}(+33.3\%)}} & \small\textit{26.2}\\
 & BFS$^{(5)}$ & 100.0 & 100.0 {\scriptsize {\color{black}(+0.0\%)}} & \small\textit{97.3} & 92.1 & {93.4} {\scriptsize {\color{red}(+1.4\%)}} & \small\textit{92.3} & {60.7} & 57.2 {\scriptsize {\color{blue}(-5.8\%)}} & \small\textit{54.5} & 33.6 & {40.6} {\scriptsize {\color{red}(+20.8\%)}} & \small\textit{39.2} & 29.1 & {33.0} {\scriptsize {\color{red}(+13.4\%)}} & \small\textit{32.0}\\
 & MCTS$^{(1)}$ & {62.2} & 59.5 {\scriptsize {\color{blue}(-4.3\%)}} & \small\textit{66.0} & 64.5 & {69.7} {\scriptsize {\color{red}(+8.1\%)}} & \small\textit{67.1} & 41.9 & {42.8} {\scriptsize {\color{red}(+2.1\%)}} & \small\textit{42.1} & 17.5 & {21.7} {\scriptsize {\color{red}(+24.0\%)}} & \small\textit{20.3} & 16.5 & {22.3} {\scriptsize {\color{red}(+35.2\%)}} & \small\textit{15.5}\\
 & MCTS$^{(10)}$ & 91.9 & {94.6} {\scriptsize {\color{red}(+2.9\%)}} & \small\textit{91.9} & 88.2 & {89.5} {\scriptsize {\color{red}(+7.4\%)}} & \small\textit{86.8} & {76.6} & 75.2 {\scriptsize {\color{blue}(-1.8\%)}} & \small\textit{75.2} & 51.7 & {54.5} {\scriptsize {\color{red}(+5.4\%)}} & \small\textit{50.3} & 32.0 & {34.0} {\scriptsize {\color{red}(+6.2\%)}} & \small\textit{29.1}\\

\bottomrule
    \end{tabular}
        }
    \caption{Full \textbf{Blocksworld} Results. CoT-Pass$^{(n)}$ denotes the rate that at least one correct solution in $n$ CoT generations. BFS$^{(b)}$ means the number of beams in BFS is $b$. MCTS$^{(n)}$ denotes $n$ iterations are performed. Number of steps is the minimum number of actions to solve this problem. A higher number of steps indicates a harder problem.}
    \label{tab:bw-res-full}
\end{table}

\subsection{CraigslistBargin Example} \label{sec:bargin-sample}
\begin{figure}[h]
    \centering
    \includegraphics{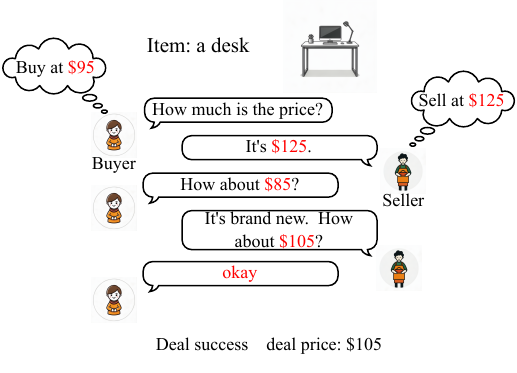}
    \caption{An example of bargaining. The profit of the seller in this deal is $\frac{105-(125+95)/2}{125-95}=-0.17$}
    \label{fig:bargin-exampl}
\end{figure}

\newpage
\subsection{Iterative Improvement with RoT} \label{sec:iter-imp}

\begin{figure}[h]
    \centering
    \begin{minipage}{0.49\linewidth\relax}
    \centering
    \begin{minipage}{0.8\linewidth\relax}
    
    \includegraphics[width=\linewidth]{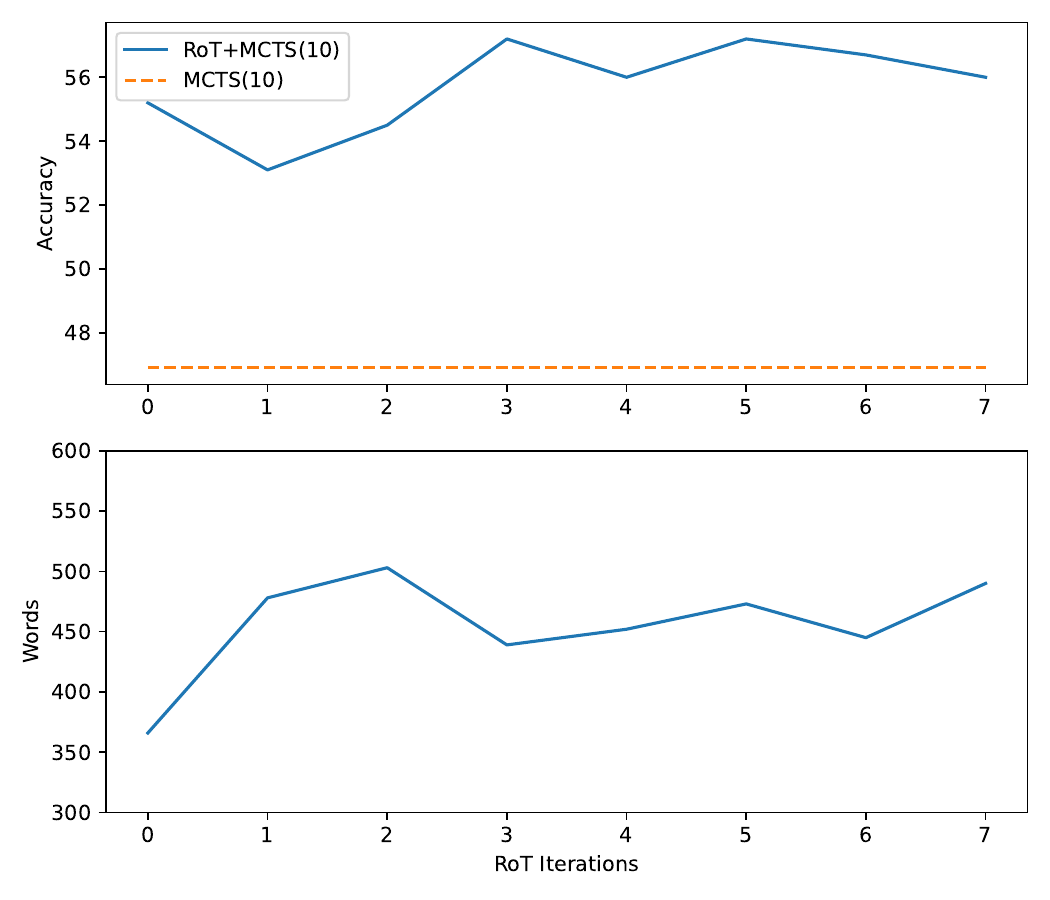}
    \caption{Word count of summarized guidelines and accurate when iteratively applying RoT to MCTS$^{(10)}$ on \textbf{Blocksworld} with step 6 with \textbf{phi-2}.}
    \label{fig:bw-iter}
    \end{minipage}
    \end{minipage}
    \begin{minipage}{0.49\linewidth\relax}
    \centering
    \begin{minipage}{0.8\linewidth\relax}
    \includegraphics[width=\linewidth]{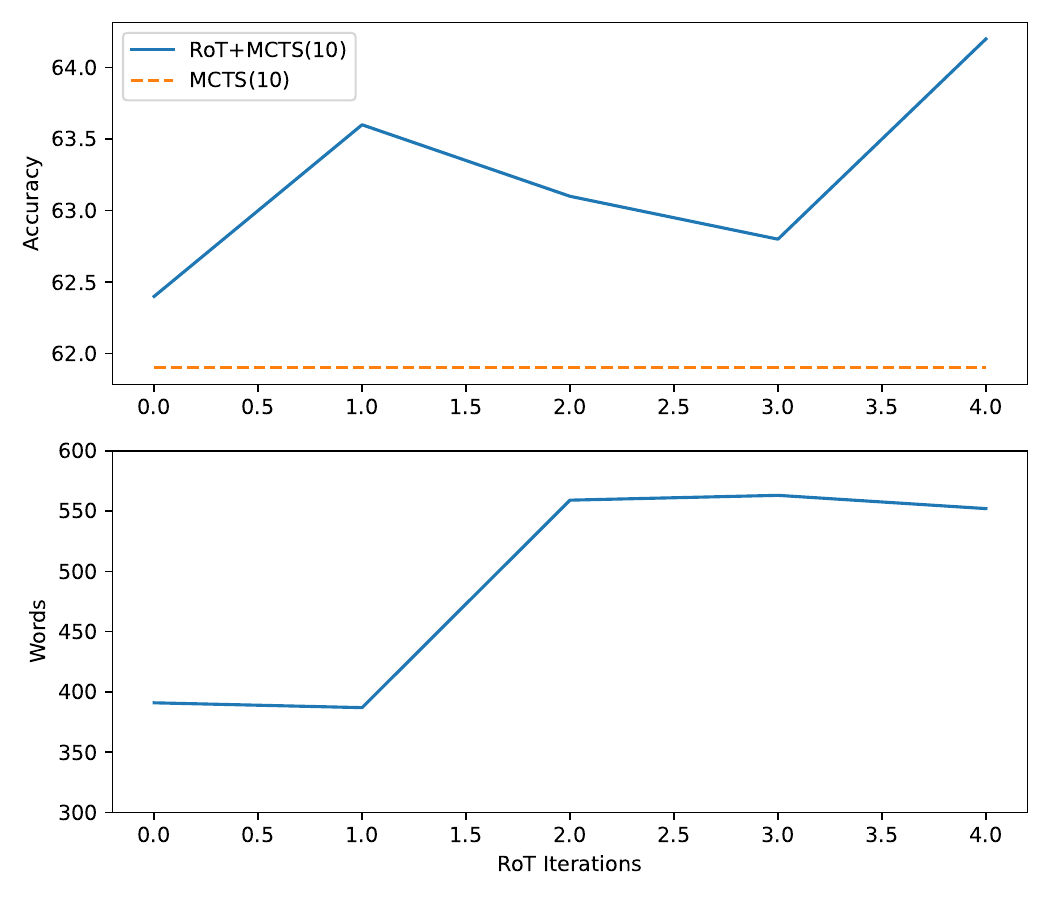}
    \caption{Word count of summarized guidelines and accurate when iteratively applying RoT to MCTS$^{(10)}$ on GSM8k with \textbf{phi-2}.}
    \label{fig:gsm8k-iter}
    \end{minipage}
    \end{minipage}
\end{figure}

We evaluate iterative improvement with RoT on the 6-step split of \textbf{Blocksworld} using MCTS$^{10}$ with \textbf{phi-2}. As shown in Figure \ref{fig:bw-iter} and Figure \ref{fig:gsm8k-iter}, the performance does not improve significantly if we iteratively perform RoT. In \textbf{Blocksword}, as the number of iterations of RoT increases, the accuracy tends to increase when the number of RoT iterations is small and soon gets saturated. We can also find that a longer guideline does not necessarily lead to better performance.

\begin{table}[h]
    \centering
    \scalebox{0.7}{
    
    \begin{tabular}{l|cccc}
    \toprule
    Method & \bf phi-2 & \bf mistral-7b & \bf mixtral-8x7b\\ \midrule
    MCTS        & 57.3      & 51.9      & 68.1 \\
    RoT-MCTS & \bf 57.7{\scriptsize{(+0.7\%)}}  & \bf 52.7{\scriptsize{(+1.5\%)}} & \bf 69.7{\scriptsize{(+2.4\%)}}  \\ \bottomrule
    \end{tabular}
    }
    \caption{AUC of \textbf{GSM8k}.}
    \label{tab:gsm8k-auc-detail}
\end{table}

\subsection{Guidelines in Prompising Decision Selection}

\begin{table*}[]
    \centering
    \scalebox{0.8}{
    \begin{tabular}{lp{11cm}c}\toprule
        Method & Guideline & Accuracy\\\midrule
        MCTS$^{(1)}$ & - & 42.4 \\\midrule
        \emph{+ problem samples} & To effectively solve math word problems, it's important to follow a structured approach. Here is a general guideline that can be applied to many types of math word problems:
        \begin{enumerate}
            \item Read the Problem Carefully: Understand what the problem is asking. Identify the key information and what the problem is asking you to find.
            \item Identify the Variables: Determine what quantities the problem is dealing with and assign symbols if necessary.
            \item Translate Words into Math: Convert the words into mathematical expressions or equations using the identified variables.
            \item Develop a Plan: Decide on the steps you need to take to solve the problem using the information given.
            \item Execute the Plan: Carry out the steps you have decided on to find the solution.
            \item Check Your Work: Verify that your solution makes sense in the context of the problem and that you have answered what was asked.
        \end{enumerate} & 44.7 \\ \midrule
        \emph{+ all states} & Comprehensive Policy for Solving Word Problems:
\begin{enumerate}
\item Initial Reading and Understanding:
\begin{itemize}
	\item Carefully read the entire word problem to fully understand the main question, the context, and all the relevant information needed for the solution.
	\item Identify initial conditions, key details, and the end goal of the problem to ensure a comprehensive understanding before beginning to solve it.
\end{itemize}
\item Breaking Down the Problem into Subquestions:
\begin{itemize}
	\item Formulate subquestions that directly relate to and build upon the main question, ensuring each subquestion is a necessary step toward reaching the final answer.
	\item Make sure all aspects of the problem are covered in the subquestions, including any conditions, contributions from different sources, rates, and sequences required for a complete solution.
	\item Avoid redundancy in subquestions and exclude any that provide unnecessary calculations or information already known.
\end{itemize}
\item Calculations and Unit Conversions:
\begin{itemize}
	\item When dealing with time conversions or solving for rates, maintain high precision by using fractions or at least three decimal places to prevent rounding errors until the final calculation.
	\item Double-check all arithmetic operations, especially additions, subtractions, and multiplications, for accuracy and ensure that they align with the broader context of the problem.
	\item In context-appropriate situations, keep answers in fractional form to represent exact figures and avoid unnecessary rounding.
\end{itemize}
\item Approach and Logical Building:
\begin{itemize}
	\item Proceed logically through the subquestions, with each providing information and answers that build upon the previous ones, culminating in the solution to the main question.
	\item Sequentially address each component of the problem, making sure not to miss any crucial step in between, such as additional money needed, contributions, or subsequent divisions.
\end{itemize}
\end{enumerate}& 44.2 \\ \midrule
    \end{tabular}
    }
\caption{Guideline samples and performance of {\bf mistral-7b} on {\bf GSM8k} when employing different selection mechanism for reflection.}
\label{tab:samples}
\end{table*}

\begin{table*}
\centering
\scalebox{0.85}{
\begin{tabular}{lp{11cm}c}\toprule
        \emph{+ random states} & This policy serves as a guideline for framing subquestions and responses in a precise and logical manner to solve complex problems, ensuring maximum accuracy and reward attainment.
\begin{enumerate}
\item
Understanding the Problem Context:
\begin{itemize}
\item Ensure you have a deep understanding of the given problem, including initial conditions, mathematical concepts required (e.g., percentages, multiplication), and the main question's objectives.
\item Accurately account for starting conditions or states, incorporating all relevant variables that influence the problem's outcome.
\end{itemize}
\item
Formulating Subquestions:
\begin{itemize}
\item Subquestions must be directly relevant and contribute meaningfully towards solving the main question. Validate this relevance carefully.
\item Keep subquestions simple, clear, and precise, using concise language to avoid confusion and misinterpretation.
\item Avoid redundant or unnecessary subquestions and eliminate any that do not bring us closer to the final solution.
\item Follow a logical progression where each subquestion builds upon previous answers, leading incrementally to the final answer.
\item Subquestions should be asked in the appropriate order that aligns with the problem-solving process.
\item Restate relevant information from previous steps when answering subquestions to prevent misunderstanding.
\end{itemize}
\item
Calculations and Computations:
\begin{itemize}
\item Carefully perform calculations required for each subquestion, double-checking arithmetic and using tools like calculators when necessary.
\item Confirm that multiplication factors and percentage increases are correctly applied; convert percentages to decimals for easier computation.
\item Accumulate values correctly according to the problem statement. For instance, apply the correct operations for totals (addition) and increases (multiplication).  
\item When integrating intermediate answers, ensure full and correct incorporation into subsequent problem-solving steps.
\end{itemize}
\item
Review and Validation:
\begin{itemize}
\item Review each calculation and logical flow at every step to verify accuracy before proceeding.
\item Synthesize all subanswers thoroughly to construct an accurate final answer to the main question.
\item Consistently check the logic and mathematical rationale behind each step to ascertain that the path from subanswers to final solutions is logical.
\item After providing responses, review the mathematical correctness and ensure that intermediate steps logically lead to the final answer.
\item Any time a reward is marked as false, carefully analyze the actions and responses that led to the error. Adjust future strategies based on this analysis.
\end{itemize}
\item
Consistency in Rewards:
\begin{itemize}
\item Maintain a consistent reward system where correct answers are indicated with the appropriate numerical value (e.g., 1.0) and incorrect answers are clearly marked with 'False' or a reduced reward.
\end{itemize}
\end{enumerate}
& 45.0\\\bottomrule
\end{tabular}}
\caption{Guideline samples and performance continue.}
\label{tab:my_label}
\end{table*}

\begin{table*}
\centering
\scalebox{0.85}{
\begin{tabular}{lp{11cm}c}\toprule
        \emph{+ prom. states ($\lambda=0.5$)} & \begin{enumerate}
\item Thorough Understanding and Problem Statement:
\begin{itemize}
\item Read the problem statement thoroughly to fully understand the context and what is specifically being asked.
\item Identify and recognize essential data that directly affects the calculation needed for the main question's answer.
\end{itemize}
\item Formulation of Subquestions:
\begin{itemize}
\item Develop relevant subquestions that address individual components of the problem and lead toward the solution, ensuring each one progresses logically toward solving the main problem.
\item Frame subquestions to provide necessary intermediate values required in later steps, without introducing extraneous information.
\item Ensure subquestions are structured to answer the main question in a logical sequence, providing clear, step-by-step reasoning within subanswers.
\end{itemize}
\item
Calculation Precision and Accuracy:
\begin{itemize}
\item Always be as precise as possible in calculations, maintaining exactness in intermediate values and avoiding premature rounding, especially when dealing with fractions or divisions.
\item When addressing percentages, use the correct mathematical operations to calculate increases or other quantities, ensuring the correct base value is used.
\item Validate subanswers by re-confirming calculations if results seem unlikely or to rectify simplistic errors, double-checking for accuracy and consistency in units.
\end{itemize}
\item
Simplification and Rounding Conventions:
\begin{itemize}
\item Provide simplified fractions that maintain the same level of accuracy as their decimal form when possible.
\item Final answers must reflect proper rounding conventions, especially involving currency, rounding off to the nearest cent for financial accuracy.
\end{itemize}
\item
Consistency and Coherence:
\begin{itemize}
\item Maintain consistency in units throughout the problem, ensuring all units are converted and used consistently.
\item Check if subanswers are logically consistent, mathematically correct, and coherent when assembled to solve the overall problem.
\end{itemize}
\item
Verification and Review:
\begin{itemize}
\item Perform a final check of the subanswer against the question to validate that it addresses the subquestion fully and revisit steps when necessary.
\item Verify the completeness of the solution, ensuring all steps are covered and no intermediate values are missing.
\item Prior to finalizing the answer, verify if it answers the main question sensibly and accurately.
\end{itemize}
\item
Learning from Rewards and Feedback:
\begin{itemize}
\item Use rewards as feedback to identify successful problem-solving strategies and reinforce approaches that yield higher accuracy.
\item Reflect on and learn from any mistakes made in previous problems, understanding why certain responses received lower rewards to improve future outcomes.
\end{itemize}
\end{enumerate}
& 45.6\\\bottomrule
\end{tabular}}
\caption{Guideline samples and performance continue.}
\label{tab:my_label}
\end{table*}

\begin{table*}
\centering
\scalebox{0.85}{
\begin{tabular}{lp{11cm}c}\toprule
        \emph{+ prom. states ($\lambda=0.1$)} & To solve math word problems effectively and minimize errors, integrate the various suggested policies into the following comprehensive approach:
\begin{enumerate}
\item Clarify the quantities involved, distinguishing between total amounts, unit costs, and multiplicative factors. Use correct mathematical operations based on these relationships.
\item When necessary, accurately convert between units (e.g., minutes to hours) using proper mathematical operations, and maintain consistency in units throughout the problem.
\item Directly translate the word problem’s conditions into mathematical equations or expressions, and apply correct mathematical operations.
\item
Address each sub-question in a logical order, one at a time, methodically performing operations and maintaining at least three decimal places of precision for operations with hourly rates or similar calculations.
\item
When dealing with ratios or relational costs, ensure the correct base amounts are used for calculations.
\item
Sequentially solve the sub-questions, constantly cross-referencing with the problem's conditions, and checking that each sub-answer is logical and consistent with the overall scenario.
\item
Refrain from rounding intermediate results to maintain accuracy, and only round the final result if necessary, ensuring the level of precision matches that of the given variables.
\item
Regularly ensure that all units and logical consistencies are maintained throughout the solution process, avoiding impossible situations (such as fractions of indivisible items).
\item
Double-check each step against the provided information and common sense narrative to avoid repetition of errors or misinterpretation.
\item
Before finalizing the answer, confirm that each part of the problem has been addressed and that the math operations have been applied correctly. Validate each sub-answer and the final answer, ensuring that they make sense and address the question accurately and fully.
\item
After combining sub-answers for the final solution, question its reasonableness in the context of the problem, verifying if quantities add up correctly and whether the overall solution is plausible.
\end{enumerate}& 47.3\\\bottomrule
\end{tabular}}
\caption{Guideline samples and performance continue.}
\label{tab:my_label}
\end{table*}

\newpage

\section{Experiment Settings}
\subsection{Hyperparameter settings.}
In Blocksworld, we provide the model with 4 in-context demonstrations when performing value estimation and next-state prediction. In GSM8k, we provide the model with 4 in-context demonstrations when performing action generation, value estimation, and response generation. In all experiments, we set the decoding to sample with temperature $t=0.7$. Due to limited computational resources, for \textbf{phi-2} models, we get the results by averaging 3 runs, and for the other models, we get the result by a single run. 

\subsection{Model Details}\label{sec:model}
\noindent\textbf{phi-2:} https://huggingface.co/microsoft/phi-2

\noindent\textbf{mistral-7b:} https://huggingface.co/mistralai/mistral-7b-v0.1

\noindent\textbf{mixtral-8x7b:} https://huggingface.co/mistralai/mixtral-8x7b-v0.1

\noindent\textbf{gpt-4:} gpt-4-1106

\noindent\textbf{chatgpt:} gpt-3.5-turbo-1106

\subsection{Computational Resources}
We run all our experiments on 4$\times$A800 with 80G memory.

\end{document}